\documentclass{vgtc}                          

\newenvironment{myitemize}{
\begin{itemize}
 \setlength{\itemsep}{1pt}
 \setlength{\parskip}{0pt}
 \setlength{\parsep}{0pt}}{\end{itemize}
}

\graphicspath{{figures/}} 

\usepackage{times}                     

\usepackage{tabu}                      
\usepackage{booktabs}                  
\usepackage{lipsum}                    
\usepackage{mwe}                       
\usepackage{multirow}
\usepackage{mathptmx}                  
\usepackage[normalem]{ulem}
\usepackage[linesnumbered,ruled]{algorithm2e}
\usepackage{amsfonts}
\usepackage{float}
\usepackage{amssymb}
\usepackage{algorithmic}
\usepackage{amsmath}

\DeclareMathOperator*{\mse}{MSE}

\onlineid{1028}

\vgtccategory{IEEE PacificVis Notes Paper}

\vgtcinsertpkg

\title{Meta-INR: Efficient Encoding of Volumetric Data via Meta-Learning Implicit Neural Representation}

\author{Maizhe Yang\thanks{e-mail: myang9@nd.edu} %
\and Kaiyuan Tang\thanks{e-mail: ktang2@nd.edu} %
\and Chaoli Wang\thanks{e-mail: chaoli.wang@nd.edu}}
\affiliation{\scriptsize University of Notre Dame}

\abstract{
Implicit neural representation (INR) has emerged as a promising solution for encoding volumetric data, offering continuous representations and seamless compatibility with the volume rendering pipeline. However, optimizing an INR network from randomly initialized parameters for each new volume is computationally inefficient, especially for large-scale time-varying or ensemble volumetric datasets where volumes share similar structural patterns but require independent training. To close this gap, we propose Meta-INR, a pretraining strategy adapted from meta-learning algorithms to learn initial INR parameters from partial observation of a volumetric dataset. Compared to training an INR from scratch, the learned initial parameters provide a strong prior that enhances INR generalizability, allowing significantly faster convergence with just a few gradient updates when adapting to a new volume and better interpretability when analyzing the parameters of the adapted INRs. We demonstrate that Meta-INR can effectively extract high-quality generalizable features that help encode unseen similar volume data across diverse datasets. Furthermore, we highlight its utility in tasks such as simulation parameter analysis and representative timestep selection. The code is available at \url{https://github.com/spacefarers/MetaINR}.
} 

\begin{document}


\vspace{-0.1in}
\firstsection{Introduction}

\maketitle
Volume data representation is a long-standing theme for scientific computing and visualization. 
Recently, researchers have explored using {\em implicit neural representation} (INR)~\cite{Sitzmann-SIREN-NeurIPS20} for volume data representation~\cite{Lu-neurcomp, Han-TVCG23, Tang-CG24, Tang-PVIS24}.
Using an architecture of {\em multilayer perception} (MLP), INR maps spatial coordinates to corresponding voxel values for volume fitting and stores the parameters of the MLP to represent the underlying volume.
In this context, INR offers three advantages. 
First, the model size of a fully connected network is usually much smaller than the original volume, implying that INR can achieve highly compressive results.
Second, INR can be naturally embedded into the volume rendering pipeline. A renderer can directly access the voxel values along a ray by inferring the trained network, eliminating the need to decode the entire volume beforehand.
Third, once trained, an INR can achieve arbitrary-scale interpolation without accessing the original data, significantly enhancing the convenience of post-hoc analysis.

Despite its effectiveness, the current INR network parameters optimized on one volume are not generalizable enough to be adapted to other unseen volumes.
When users represent new simulation volume data using INR, the parameters of the trained INR cannot be directly reused to accelerate training on other volumes, which leads to inefficient encoding for time-varying or ensemble datasets.
Inspired by {\em meta-learning} in neural networks~\cite{Hospedales-TPMAI22}, we propose Meta-INR, a pretraining strategy for INR that only uses partial observation of a volumetric dataset but enables rapid adaptation to unseen volumes with similar structures in a few gradient steps. 

The contributions of our work are as follows. 
First, we are the first to apply meta-learning techniques to volume data representation. 
We demonstrate that Meta-INR, pretrained using no more than 1\% of data samples, can adapt to unseen similar volumes efficiently. 
Our method reduces the computational overhead of encoding new volumes by leveraging meta-learned priors, leading to significantly faster encoding than training from scratch. 
Second, we show that the parameters of adapted INRs exhibit enhanced interpretability, making them useful for simulation parameter analysis and representative timestep selection tasks.
Finally, we evaluate Meta-INR on various datasets by comparing its performance with other training strategies.

\vspace{-0.075in}
\section{Related Work}
\vspace{-0.025in}

{\bf INR for scientific visualization.}
Using deep learning techniques for data representation~\cite{Wang-DL4SciVis} has been extensively studied recently. 
One solution uses INR, which inputs spatial coordinates and outputs corresponding voxel values to achieve the generation and reduction of scientific data. 
Typically, INR leverages the MLP architecture to represent volumetric data.
For example, Lu et al.\ \cite{Lu-neurcomp} compressed a single scalar field by optimizing an INR with weight quantization.
Han and Wang~\cite{Han-TVCG23} proposed CoordNet, a coordinate-based network to tackle diverse data and visualization generation tasks.
Tang and Wang~\cite{Tang-CG24} presented STSR-INR, leveraging an INR to generate simultaneous spatiotemporal super-resolution for time-varying multivariate volumetric data.
Han et al.\ \cite{Han-KD-INR} proposed KD-INR to handle large-scale time-varying volumetric data compression when volumes are only sequentially accessible during training.
Tang and Wang~\cite{Tang-PVIS24} designed ECNR to achieve efficient time-varying data compression by combining INR with the Laplacian pyramid for multiscale fitting.
Li and Shen~\cite{Li-TVCG24} leveraged Gaussian distribution to model the probability distribution of an INR network to achieve efficient isosurface extraction.

Besides the vanilla MLP architecture, multiple works integrate grid parameters into INR to achieve efficient encoding and rendering.
Weiss et al.\ \cite{Weiss-CGF22} implemented fV-SRN, achieving significant rendering speed gain over~\cite{Lu-neurcomp} using a volumetric latent grid.
Wurster et al.\ \cite{Wurster-TVCG24} proposed APMGSRN, which uses multiple spatially adaptive feature grids to represent a large volume.
Xiong et al.\ \cite{Xiong-TVCG24} designed MDSRN to simultaneously reconstruct the data and assess the reconstruction quality in one INR network.
Tang and Wang~\cite{Tang-VIS24} developed StyleRF-VolVis, leveraging a grid-based encoding INR to represent a volume rendering scene that supports various editings.
Yao et al.\ \cite{Yao-PVIS25} proposed ViSNeRF, utilizing a multidimensional INR representation for visualization synthesis of dynamic scenes, including changes of transfer functions, isovalues, timesteps, or simulation parameters. 
Gu et al.\ \cite{Gu-CG23} and Lu et al.\ \cite{YF-Lu-VISSP24} presented NeRVI and FCNR, respectively, utilizing INRs for the effective compression of a large collection of visualization images. 
Unlike existing works that mainly focus on network architecture design, this paper aims to develop a pretraining strategy for optimizing the initial parameters of an INR network to enhance the representation generalizability.

{\bf Meta-learning.}
Meta-learning is a deep learning technique primarily aiming for few-shot learning~\cite{Song-ACM23}.
Numerous recent studies have explored using it to optimize the initialization of neural networks, enabling them to adapt to new tasks with just a few steps of gradient descent.
For instance, 
Sitzmann et al.\ \cite{Sitzmann-MetaSDF-NeurIPS20} leveraged meta-learners to generalize INR across shapes.
Tancik et al.\ \cite{Tancik-CVPR21} employed meta-learning to initialize INR network parameters according to the underlying class of represented signals.
Emilien et al.\ \cite{Emilien-TMLR} proposed COIN++, a neural compression framework that supports encoding various data modalities with a meta-learned base network.
Similar to these works, we develop Meta-INR based on existing meta-learning algorithms~\cite{Finn-ICML17, Nichol-arXiv18}.
However, we focus on applying meta-learning in INR for volume data representation, setting it apart from existing studies.

\vspace{-0.075in}
\section{Meta-INR}
\vspace{-0.025in}

Meta-INR is a pretraining approach designed to optimize the initial parameters of an INR network for efficient finetuning on unseen volumes.
The training pipeline of Meta-INR consists of two sequential stages: {\em meta-pretraining} and {\em volume-specific finetuning}. 
\begin{myitemize}
\vspace{-0.05in}
    \item The {\em meta-pretraining} stage optimizes a meta-model on a sparse subsampled volumetric dataset, utilizing less than 1\% of the original data, to learn the initial parameters of an INR network that supports rapid adaptation to other unseen volumes within the dataset. 
    \item The {\em volume-specific finetuning} stage finetunes the initial parameters of the meta-model on a specific volume, resulting in a volume-specific adapted INR for high-fidelity volume reconstruction.
\vspace{-0.05in}
\end{myitemize}
The adapted INR shares the same network architecture, $\Phi: \mathbb{R}^3 \rightarrow \mathbb{R}$ as the meta-model, which maps a 3D coordinate to the corresponding voxel value.
Let $\theta_{m}$ denote the parameters of the meta-model.
For a volumetric dataset $\mathbf{D} = \{ d_1, d_2, \dots, d_i, \dots, d_T \}$ that contains $T$ timesteps or ensembles, one volume $d_i$ consists of a set of coordinate-value pairs $(\mathbf{C}_i, \mathbf{V}_i)$, where $\mathbf{C} = \{ (x_1, y_1, z_1), (x_2, y_1, z_1), \dots \}$ is a set of spatial coordinates and $\mathbf{V} = \{ v_1, v_2, \dots \}$ is the corresponding voxel values at these positions.
After meta-pretraining, we finetune $\theta_{m}$ on each set of coordinate-value pairs in $\mathbf{D}$ independently, resulting in a series of volume-specific adapted INRs with parameters $\{ \theta_{1}, \theta_{2}, \dots, \theta_{T} \}$ that can represent each volume within the temporal or ensemble sequence.

\begin{algorithm}[htb]
    \caption{Meta-Pretraining}
    \label{alg:meta-pretraining}
        \KwIn{Volumetric dataset $\mathbf{D}$ with $T$ timesteps or ensembles, spatial and temporal downsampling intervals $\lambda_s$ and $\lambda_t$, inner and outer loop learning rates $\alpha$ and $\beta$, number of inner-loop steps $K$.}
        \KwOut{Optimized meta-model parameters $\theta_{m}$.}
        Subsample $\mathbf{D}$ under $\lambda_s$ and $\lambda_t$ to obtain downsampled dataset $\hat{\mathbf{D}} = \{ \hat{d}_1, \dots, \hat{d}_{T'} \}$ with $T'$ timesteps or ensembles
        
        Randomly initialize meta-model parameters $\theta_{m}$
        
        \While{not done}{
            Initialize gradients: $\nabla \theta_{m} = 0$\;
            \For{all $\hat{d}_i$}{
                Clone parameters: $\theta' \leftarrow \theta_{m}$\;
                Randomly sample batches of coordinate-value pairs $(\mathbf{C}_i, \mathbf{V}_i)$ from $\hat{d}_i$\;
                \For{$k = 1 \text{ to } K$}{
                    Compute loss: $\mathcal{L} \leftarrow \mse(\Phi(\mathbf{C}_i; \theta'), \mathbf{V}_i)$\;
                    $\theta' \leftarrow \theta' - \alpha \nabla_{\theta'} \mathcal{L}$\;
                }
                $\nabla \theta_{m} \leftarrow \nabla \theta_{m} + (\theta - \theta')$\;
            }
            Update meta-model parameters: $\theta_{m} \leftarrow \theta_{m} - \beta \frac{\nabla \theta_{m}}{T'}$\;
        }
        \Return $\theta_{m}$
\end{algorithm}

\vspace{-0.05in}
\subsection{Meta-Pretraining}
\label{subsec:mp}

The meta-pretraining stage optimizes the parameters of $\theta_{m}$, serving as the initial parameters in the volume-specific finetuning stage. 
Meta-pretraining is data efficient, requiring only a small portion of the data to derive sufficiently generalizable $\theta_{m}$.
In this paper, we achieve this by spatiotemporally subsampling the original dataset $\mathbf{D}$.
Specifically, we subsample $\mathbf{D}$ using an interval of $\lambda_s$ on each spatial dimension and $\lambda_t$ on the temporal dimension, resulting in a downsampled dataset $\hat{\mathbf{D}}$.

The spatial subsampling interval $\lambda_s=4$ and temporal subsampling interval $\lambda_t=2$ are chosen empirically, balancing pretraining efficiency and quality of the prior. This configuration retains structural patterns for most datasets without significant accuracy loss.
It further implies that only $\frac{1}{(4^3\times2)}\times100\%\thickapprox0.78\%$ original data samples are used for pretraining.

Then, we consider optimizing $\theta_{m}$ as a meta-learning problem and propose to leverage a MAML-like algorithm~\cite{Finn-ICML17}.
In particular, we randomly initialize the meta-model parameters and iteratively update $\theta_{m}$ through a nested inner and outer loop. In the inner loop, for each subsampled volumetric dataset $\hat{d}_i$ in $\hat{\mathbf{D}}$, we finetune a cloned set of parameters $\theta'$ using $K$ gradient steps on randomly sampled batches of coordinate-value pairs $(\mathbf{C}_i, \mathbf{V}_i)$. The inner-loop loss is computed as the mean squared error (MSE) between the predicted and ground-truth (GT) voxel values. After processing all volumes in $\hat{\mathbf{D}}$, the outer loop accumulates the gradients from the inner loop and updates $\theta_{m}$ using the average gradient across the volumes. The optimization continues until $\theta_{m}$ converges.

\vspace{-0.05in}
\subsection{Volume-Specific Finetuning}
\label{subsec:vsf}

Once the meta-pretraining stage finishes, we utilize $\theta_{m}$ as initial parameters to finetune each adapted INR on a specific volume in $\mathbf{D}$.
The finetuning process mirrors the inner-loop adaptation during meta-pretraining, where the pre-trained initial parameters $\theta_{m}$ are taken and updated in $K$ gradient steps. 
The only difference is that we utilize all available data points in this stage instead of a subsampled version of the volumetric dataset.
This ensures that the finetuned parameters follow a more accurate gradient descent direction from $\theta_{m}$, leading to improved reconstruction accuracy.
After the volume-specific finetuning stage, we can obtain a series of volume-wise adapted INRs, each representing a specific volume within the target time-varying or ensemble sequence.

\begin{table}[htb]
\caption{Experimented datasets and their respective settings.}
\centering
{\scriptsize
\begin{tabular}{c|ccc}
  & dimension & \# of timesteps     \\ 
 dataset & (x$\times$y$\times$z) & or ensembles    \\ \hline
 earthquake &  256$\times$256$\times$96  &  598\\ 
 half-cylinder~\cite{Rojo-TVCG19} &  640$\times$240$\times$80  &  20\\ 
 ionization~\cite{Whalen-TAJ08} & 600$\times$248$\times$248  &    30\\
 Tangaroa~\cite{Popinet-JAOT04} & 300$\times$180$\times$120 &   30\\
 vortex~\cite{silver1997tracking} & 128$\times$128$\times$128 &   15 \\
 Nyx~\cite{Almgren-AJ13}        & 256$\times$256$\times$256 & 209
\end{tabular}
}
\label{tab:datasets}
\end{table}
 
\begin{table}[!htb]
\caption{Average PSNR (dB), LPIPS, CD values across all timesteps, and overall encoding time (ET), pretraining time (PT), and total time (TT) for encoding different time-varying datasets. ``p.t." stands for pretrained. The chosen isovalues for computing CD are $0.0$, $-0.6$, $-0.8$, and $-0.2$, respectively. `s', `m', and `h' denote seconds, minutes, and hours. The best metric values are shown in bold.}
\centering
{\scriptsize
\resizebox{\columnwidth}{!}{
\begin{tabular}{c|c|ccc|ccc}
dataset		&method & PSNR$\uparrow$ & LPIPS$\downarrow$ & CD$\downarrow$& ET &  PT & TT\\ \hline
\multirow{3}{*}{\shortstack{half-cylinder}}
&SIREN& 48.89&	0.0729& 0.5081& 4h 28m& --- &4h 28m\\ 
&p.t.\ SIREN& 35.87&	0.1113& 2.6655& 42m 19s&2h 39m&3h 21m\\ 
&Meta-INR& \textbf{55.92}&	\textbf{0.0705}& \textbf{0.2989}& 43m 45s&2h 14m&2h 58m\\  \hline
\multirow{3}{*}{\shortstack{ionitzation}}
&SIREN& 48.43&	0.0591& 0.5534& 21h 7m& --- &21h 7m\\ 
&p.t.\ SIREN& 40.67&	0.1189& 1.6666& 3h 21m&11h 51m&15h 12m\\ 
&Meta-INR& \textbf{51.30}&	\textbf{0.0576}&\textbf{ 0.4622}& 3h 15m&11h 28m&14h 43m\\ \hline
\multirow{3}{*}{\shortstack{Tangaroa}}
&SIREN& 43.22&	0.1989& 1.3144& 3h 29m& --- &3h 29m\\ 
&p.t.\ SIREN& 40.46&	0.3387& 16.075& 33m 23s&2h 10m &2h 43m\\ 
&Meta-INR& \textbf{48.68}&	\textbf{0.1969}&\textbf{ 0.5499}& 33m 59s&2h 7m &2h 41m\\ \hline
\multirow{3}{*}{\shortstack{vortex}}
&SIREN& 46.06&	0.0358& 0.2484& 24m 22s& --- &24m 22s\\ 
&p.t.\ SIREN& 26.22&	0.2541& 2.0146& 5m 17s&22m 49s &28m 06s\\ 
&Meta-INR& \textbf{48.19}&	\textbf{0.0294}& \textbf{0.2029}& 5m 10s&18m 44s &23m 54s\\ 
\end{tabular}
}
}
\label{tab:baseline-metrics}
\end{table}

\begin{figure}[!htb]
 \begin{center}
 $\begin{array}{c@{\hspace{0.025in}}c@{\hspace{0.025in}}c@{\hspace{0.025in}}c}
 \includegraphics[width=0.23\linewidth]{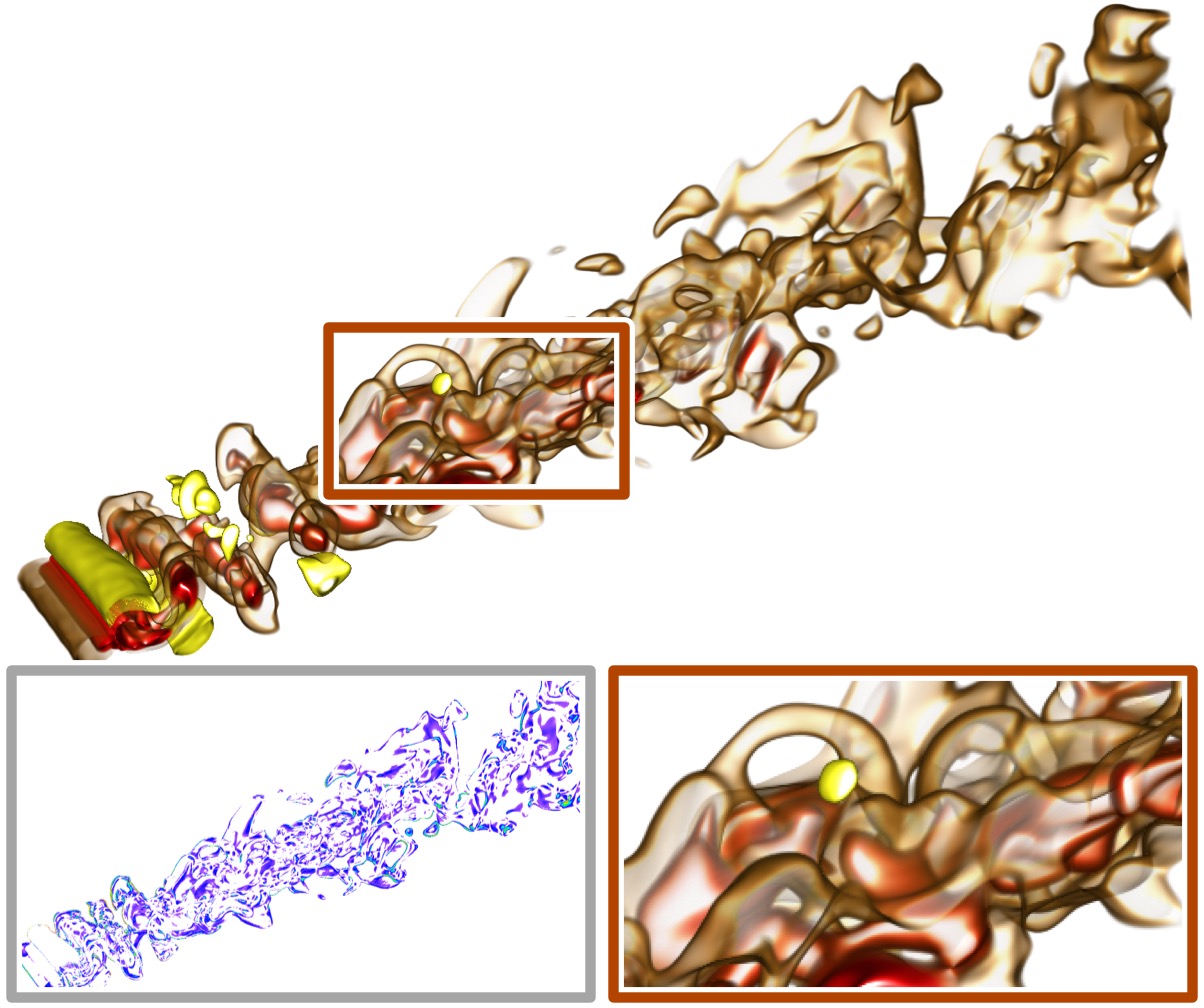}&
 \includegraphics[width=0.23\linewidth]{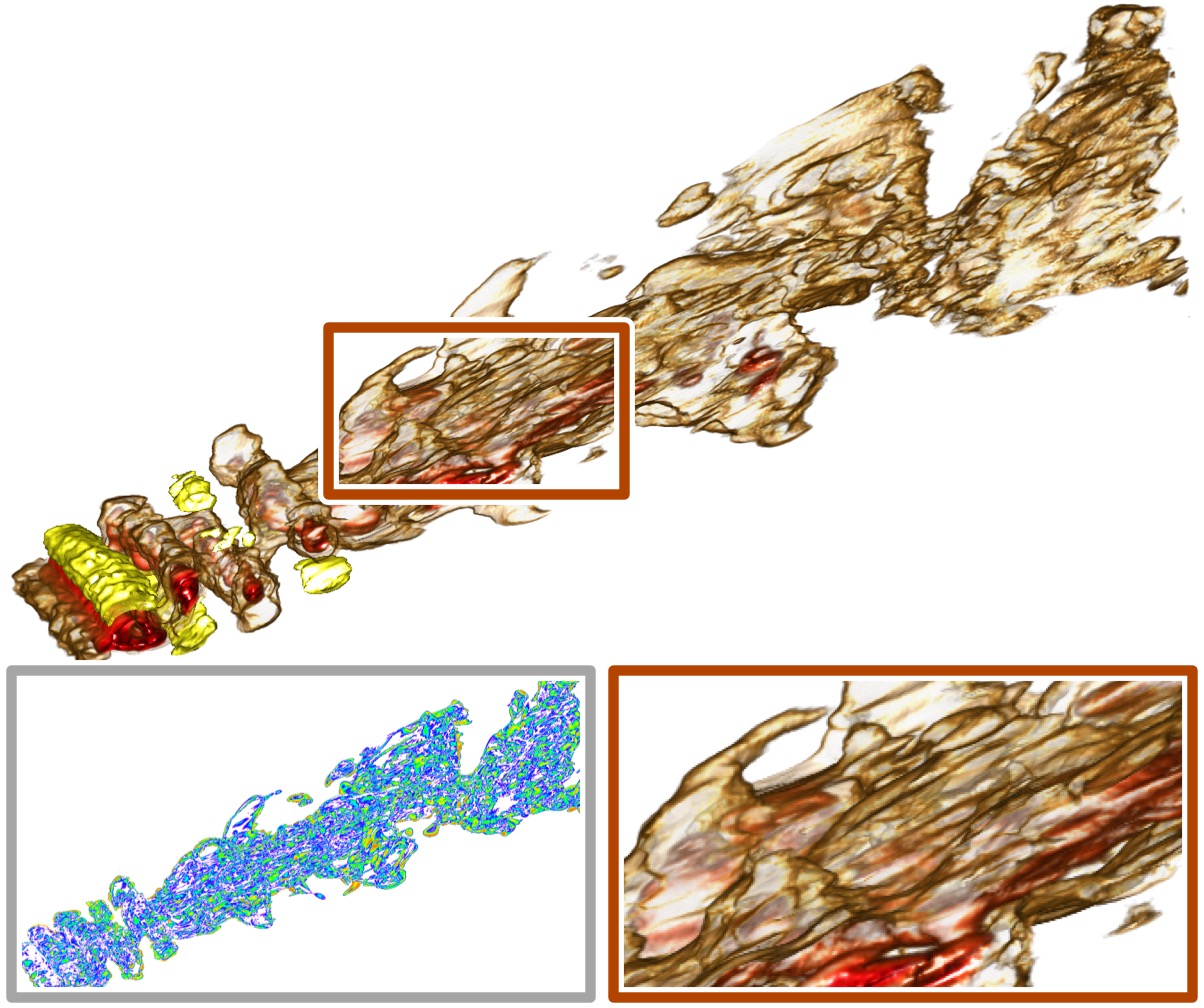}&
  \includegraphics[width=0.23\linewidth]{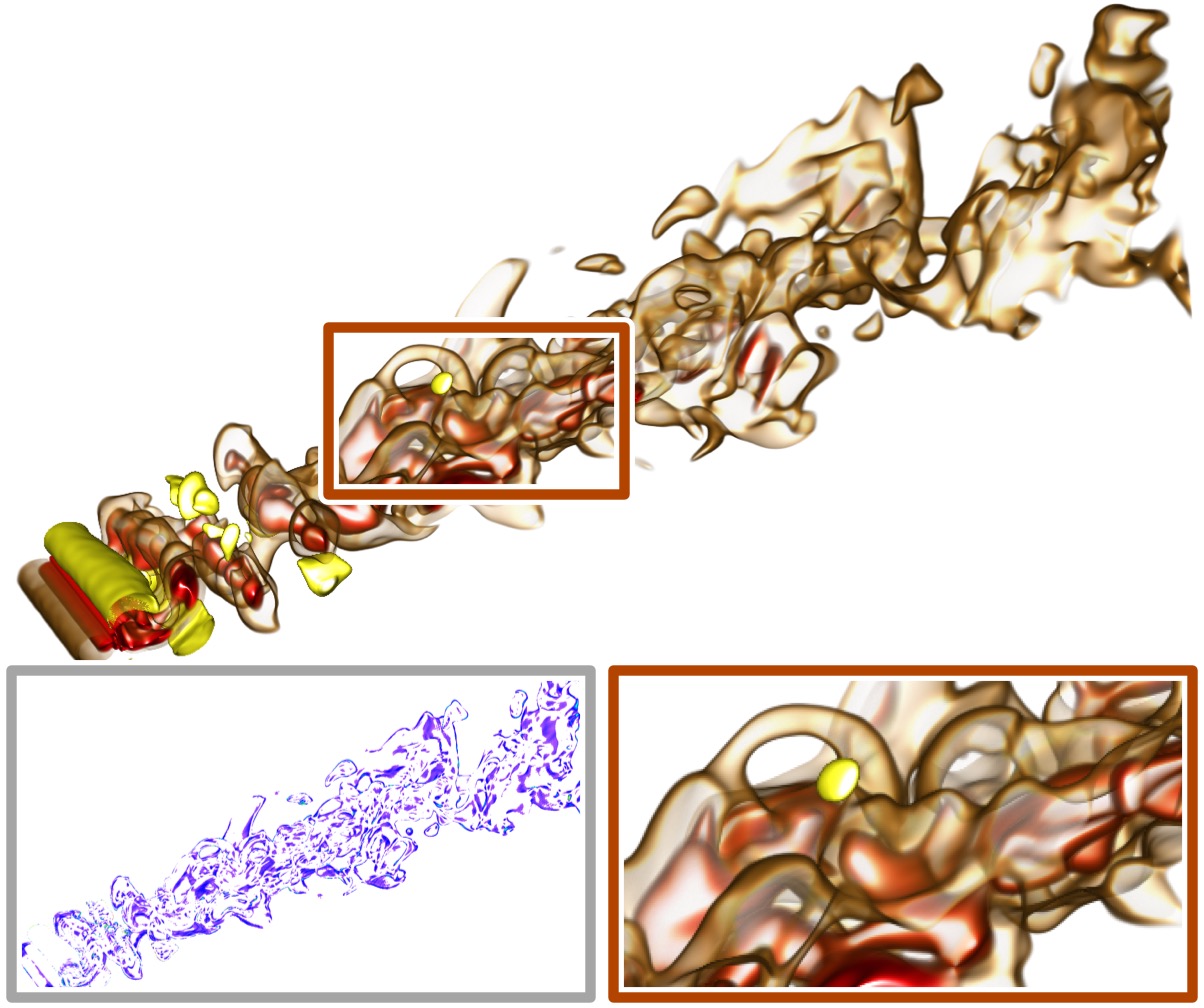}&
 \includegraphics[width=0.23\linewidth]{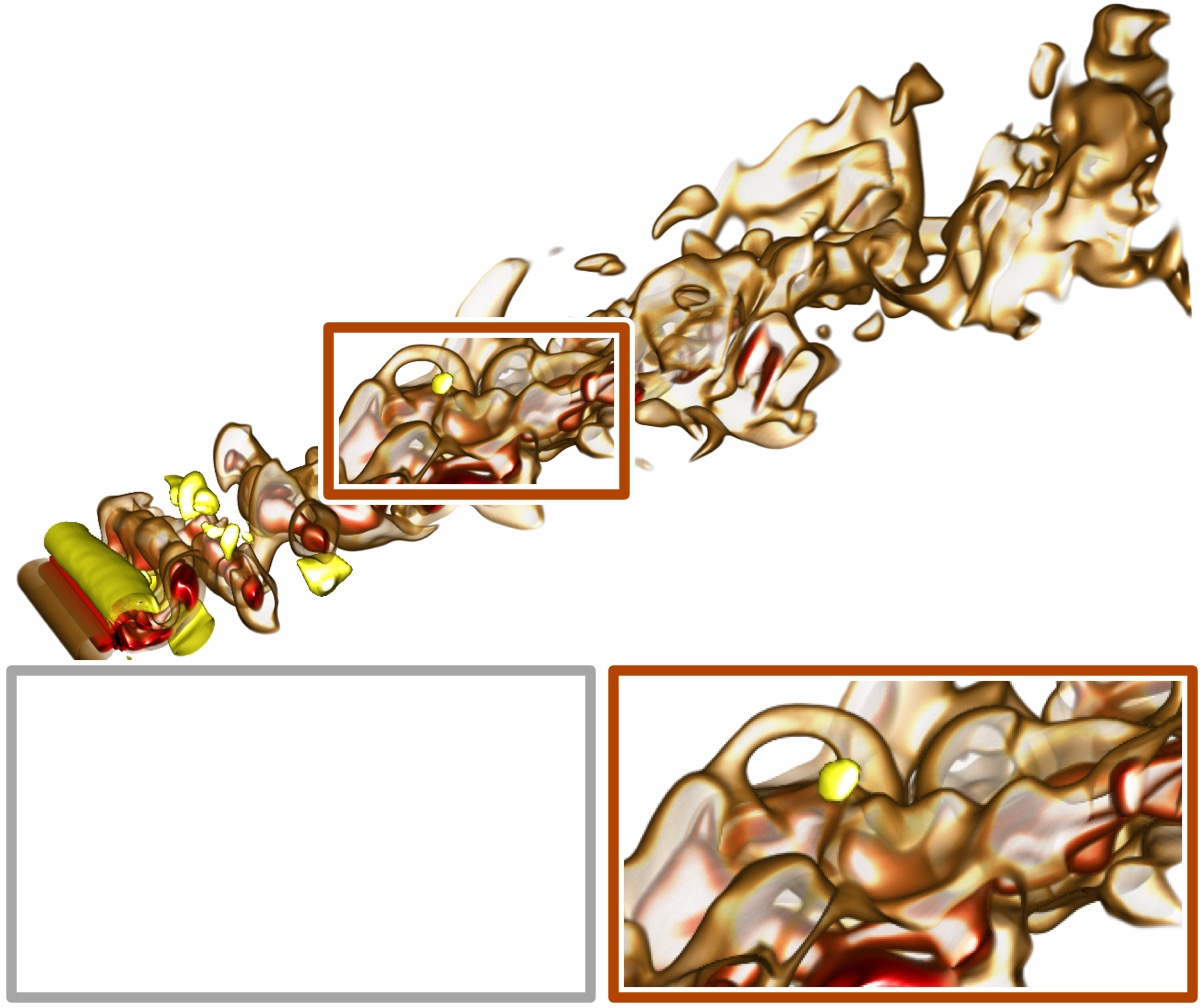} \\
 \includegraphics[width=0.23\linewidth]{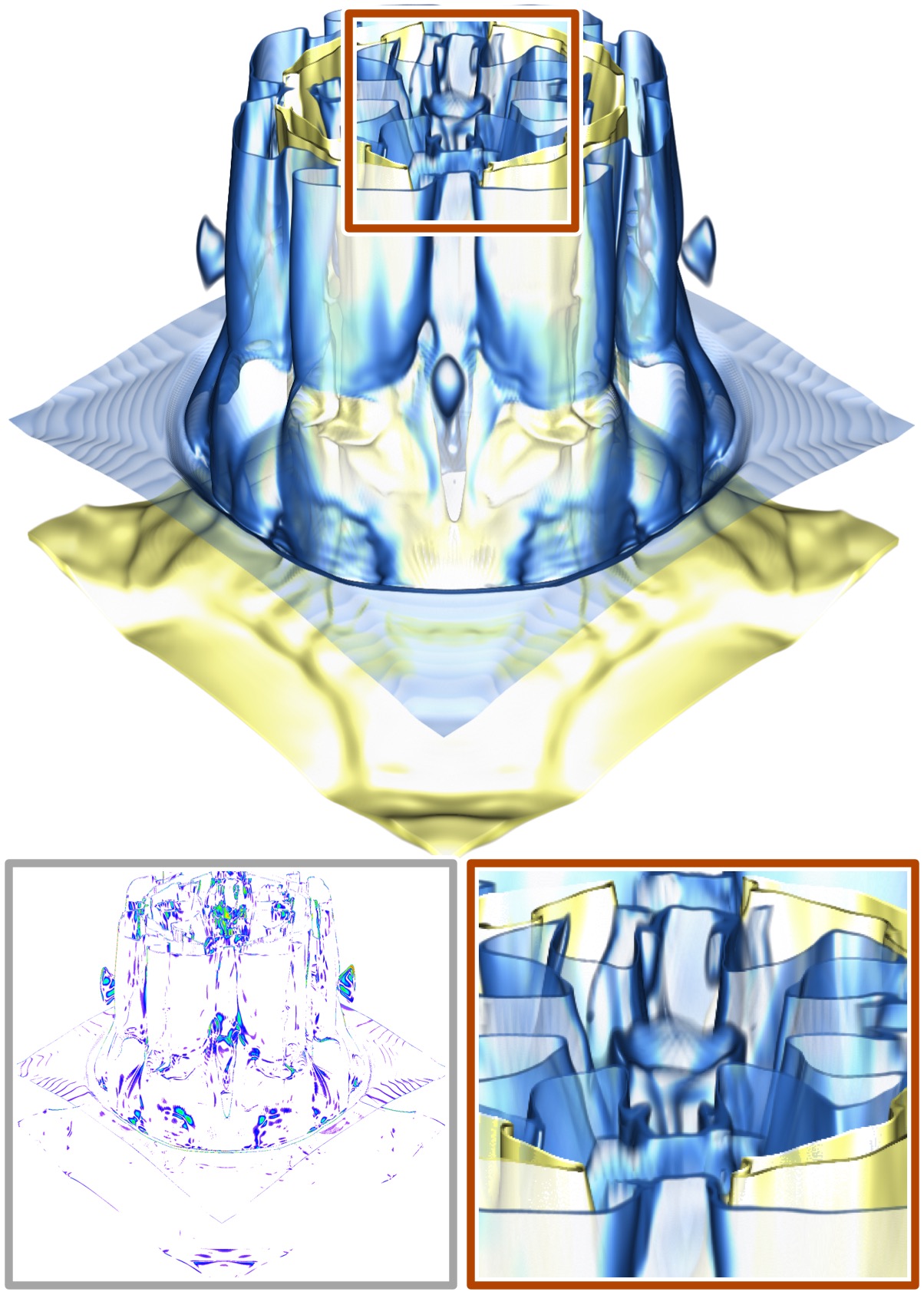}&
 \includegraphics[width=0.23\linewidth]{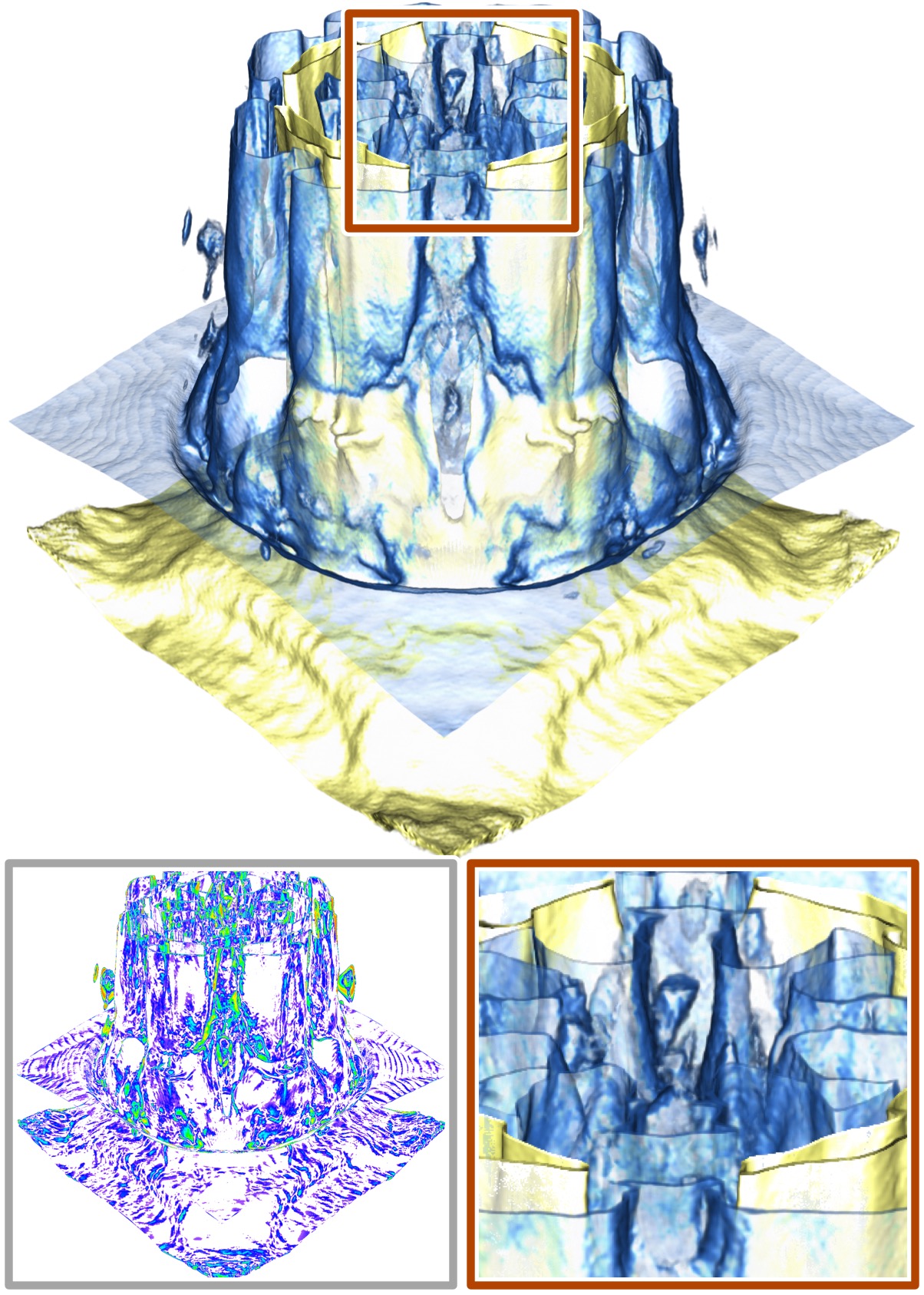}&
  \includegraphics[width=0.23\linewidth]{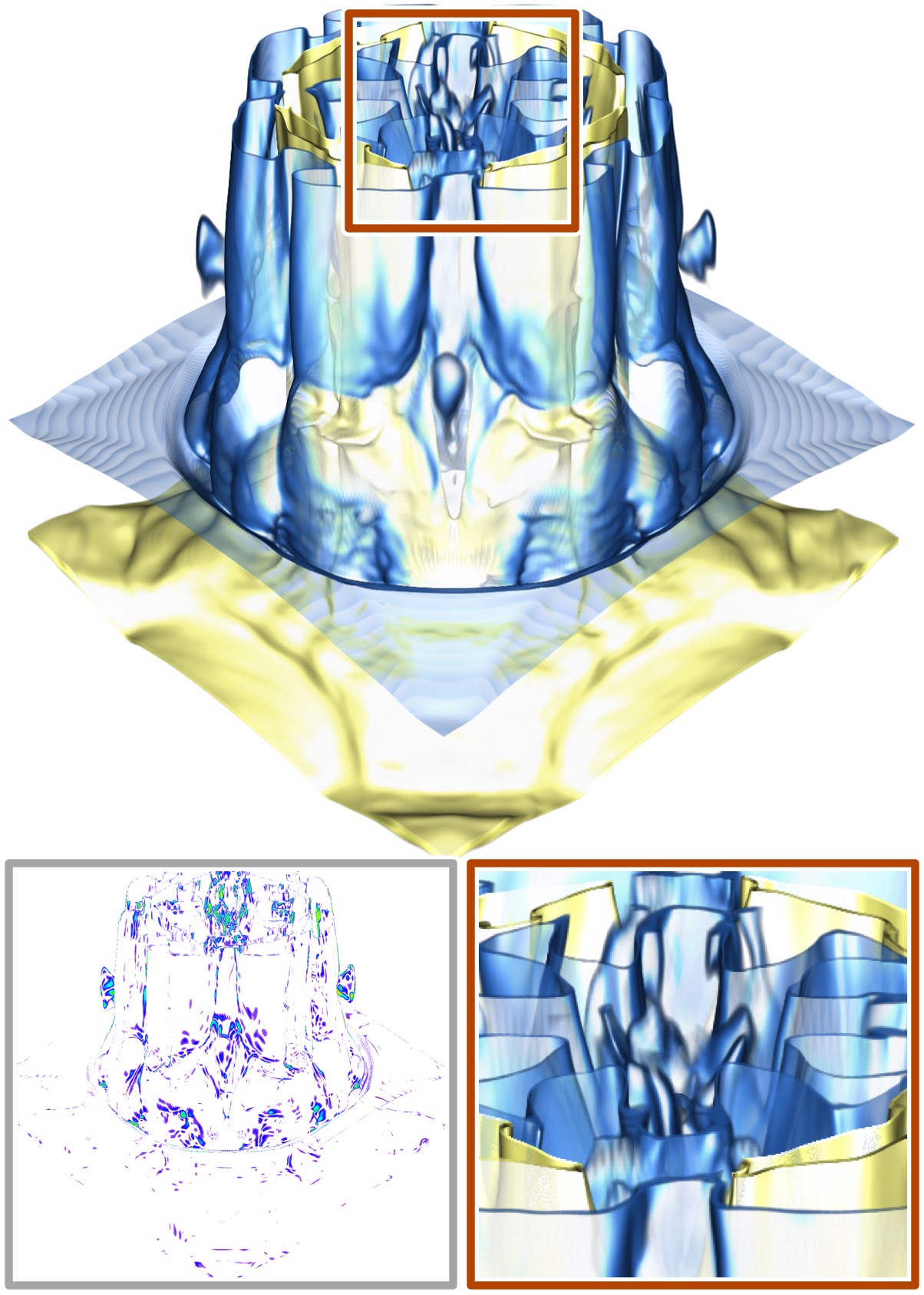}&
 \includegraphics[width=0.23\linewidth]{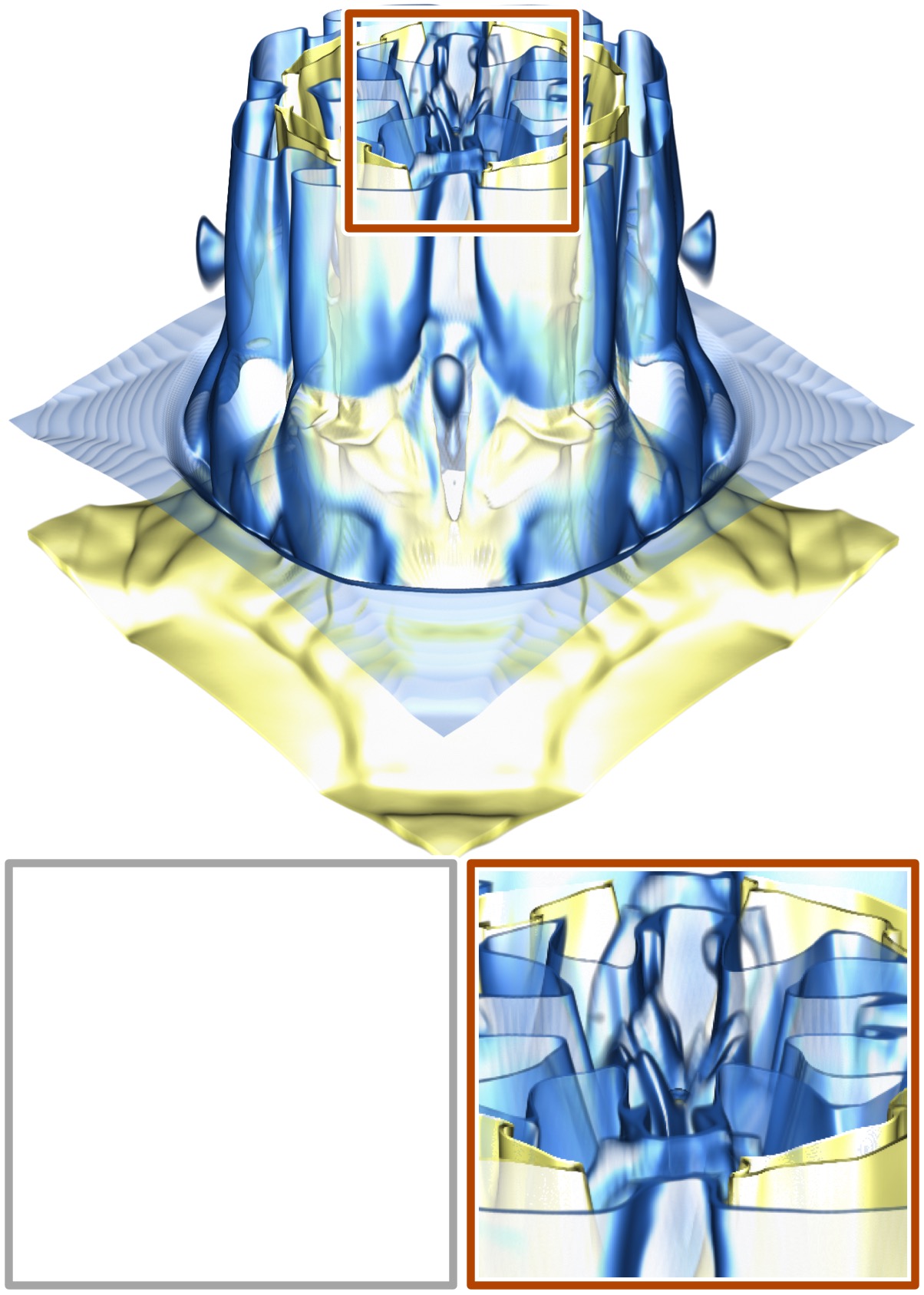} \\
 \includegraphics[width=0.23\linewidth]{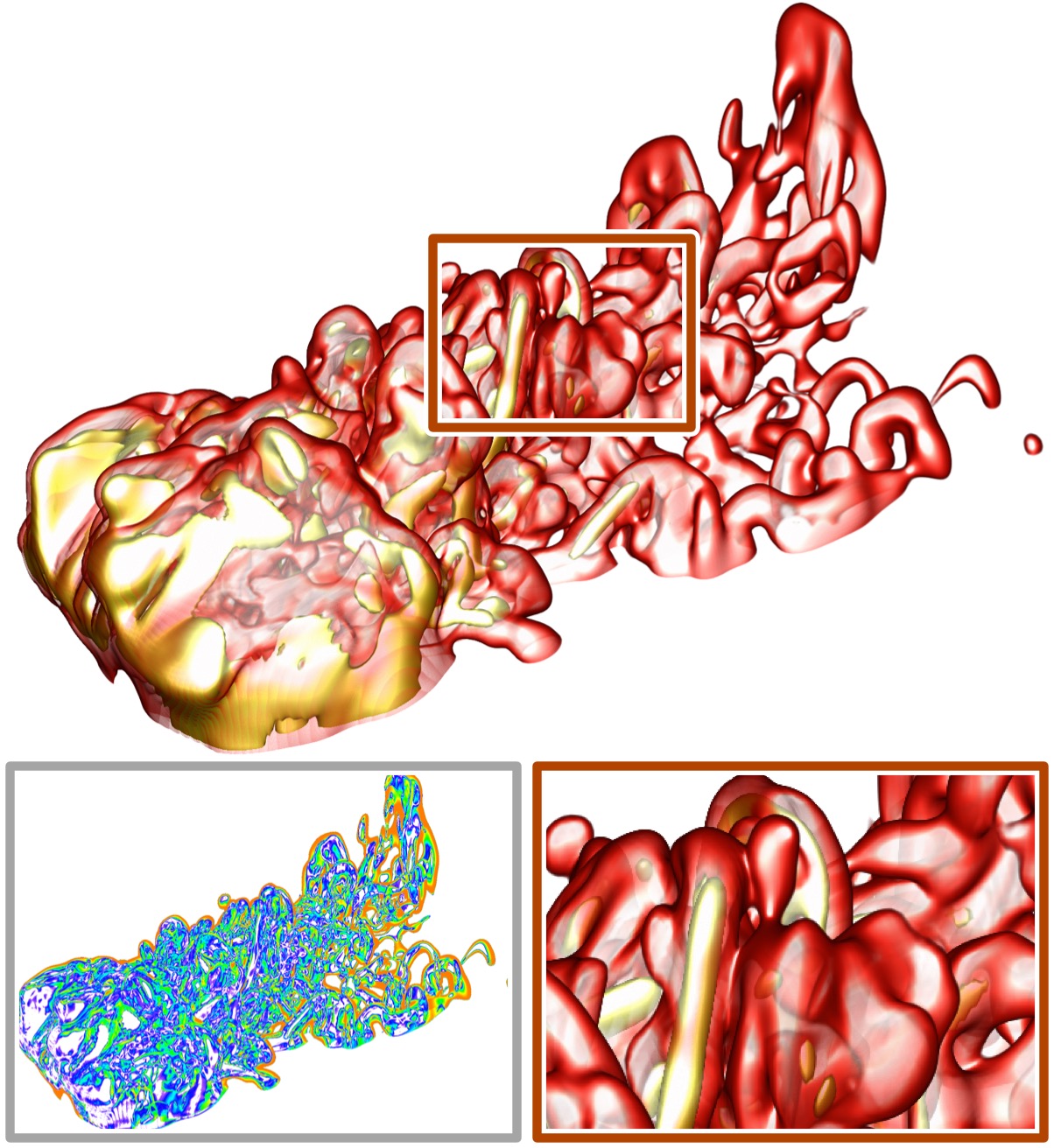}&
 \includegraphics[width=0.23\linewidth]{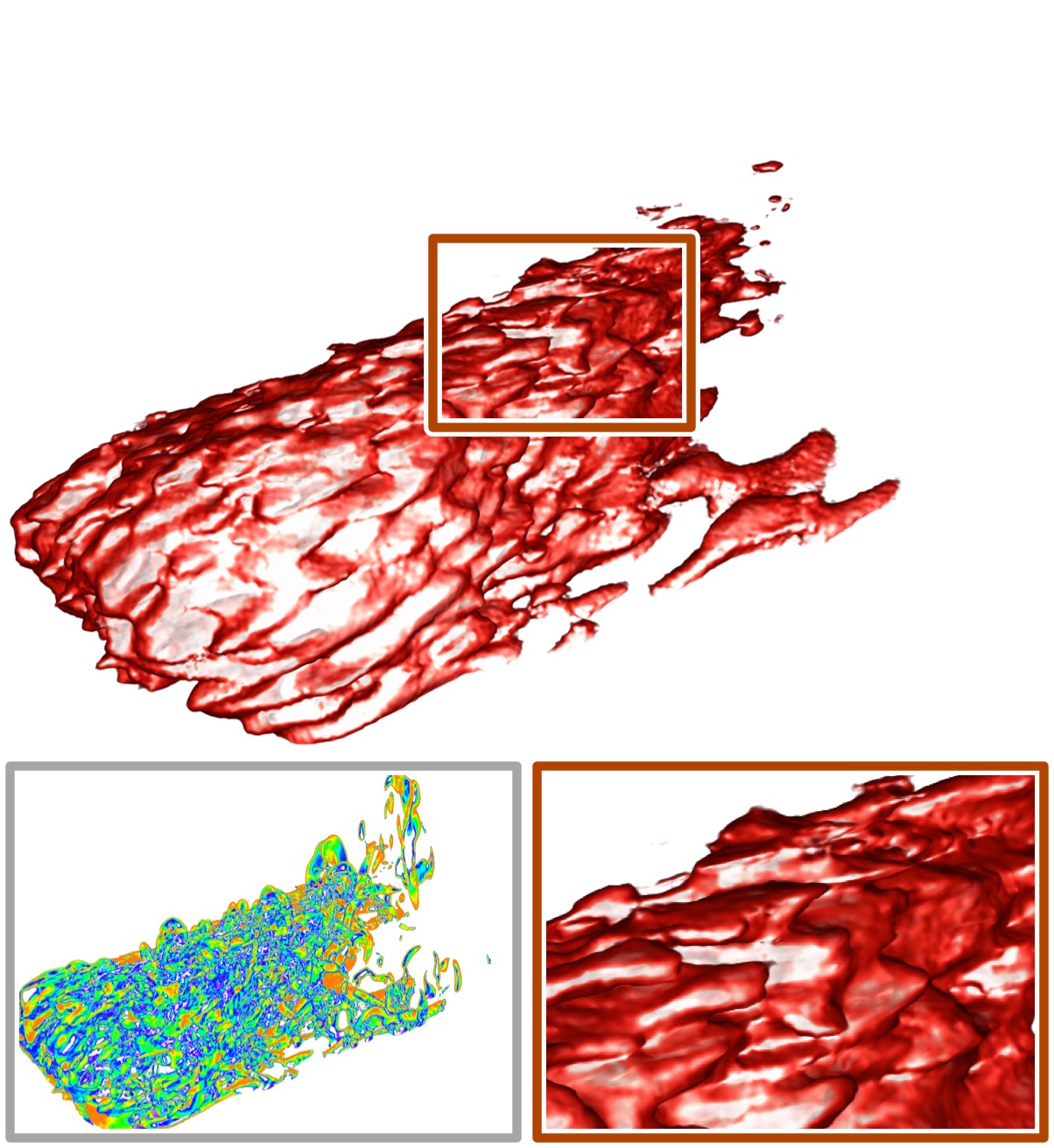}&
  \includegraphics[width=0.23\linewidth]{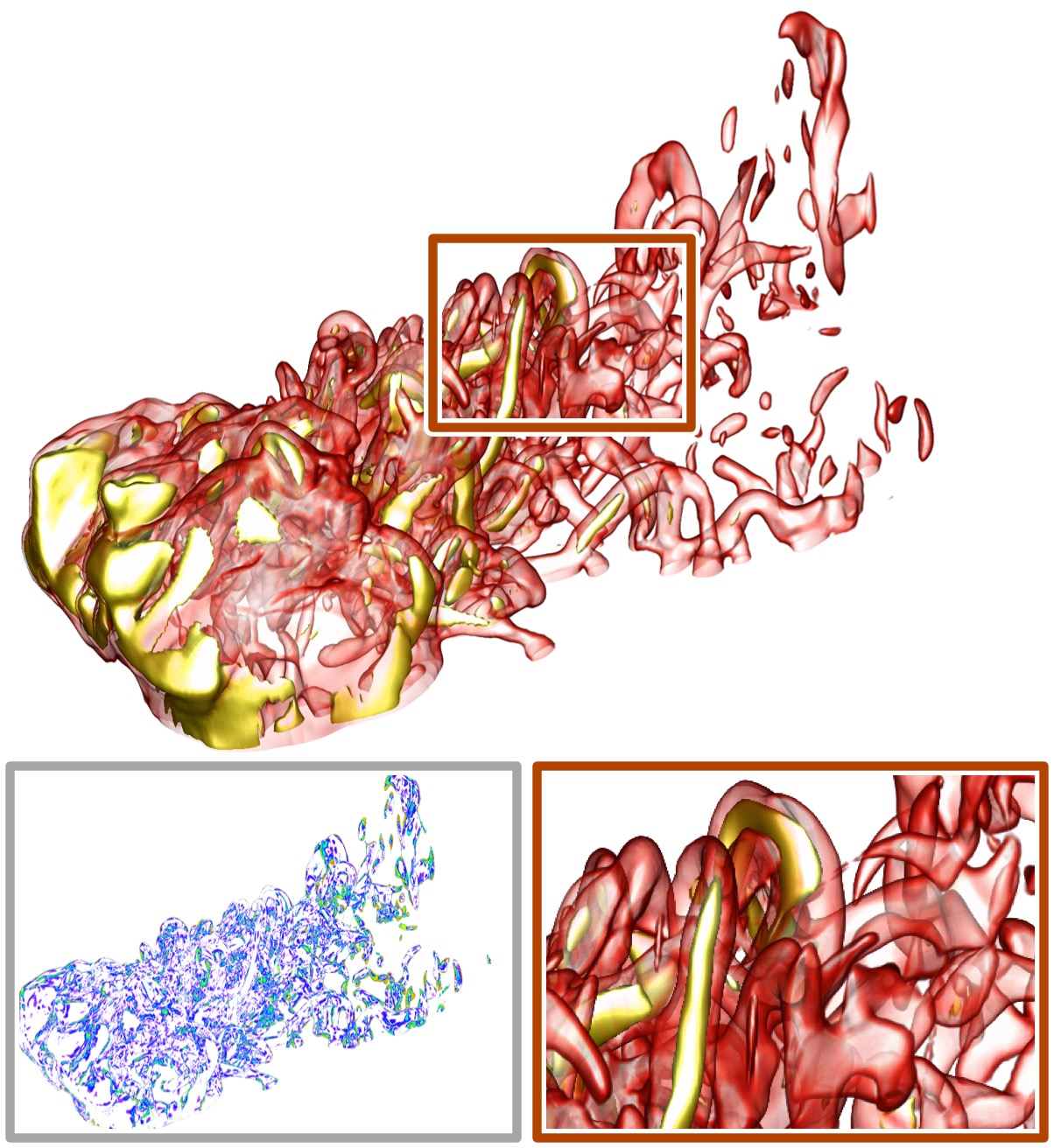}&
 \includegraphics[width=0.23\linewidth]{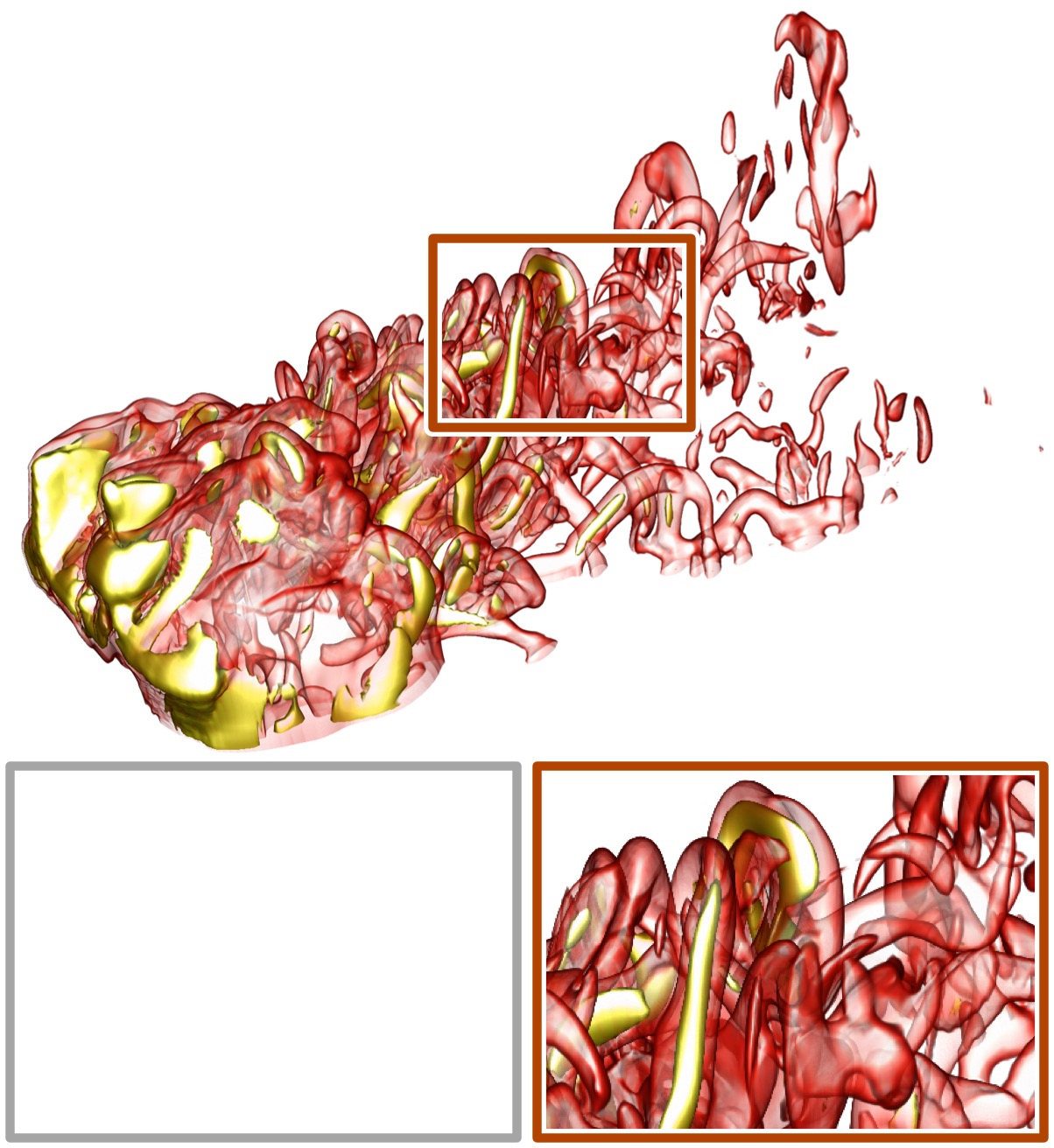} \\
 \includegraphics[width=0.23\linewidth]{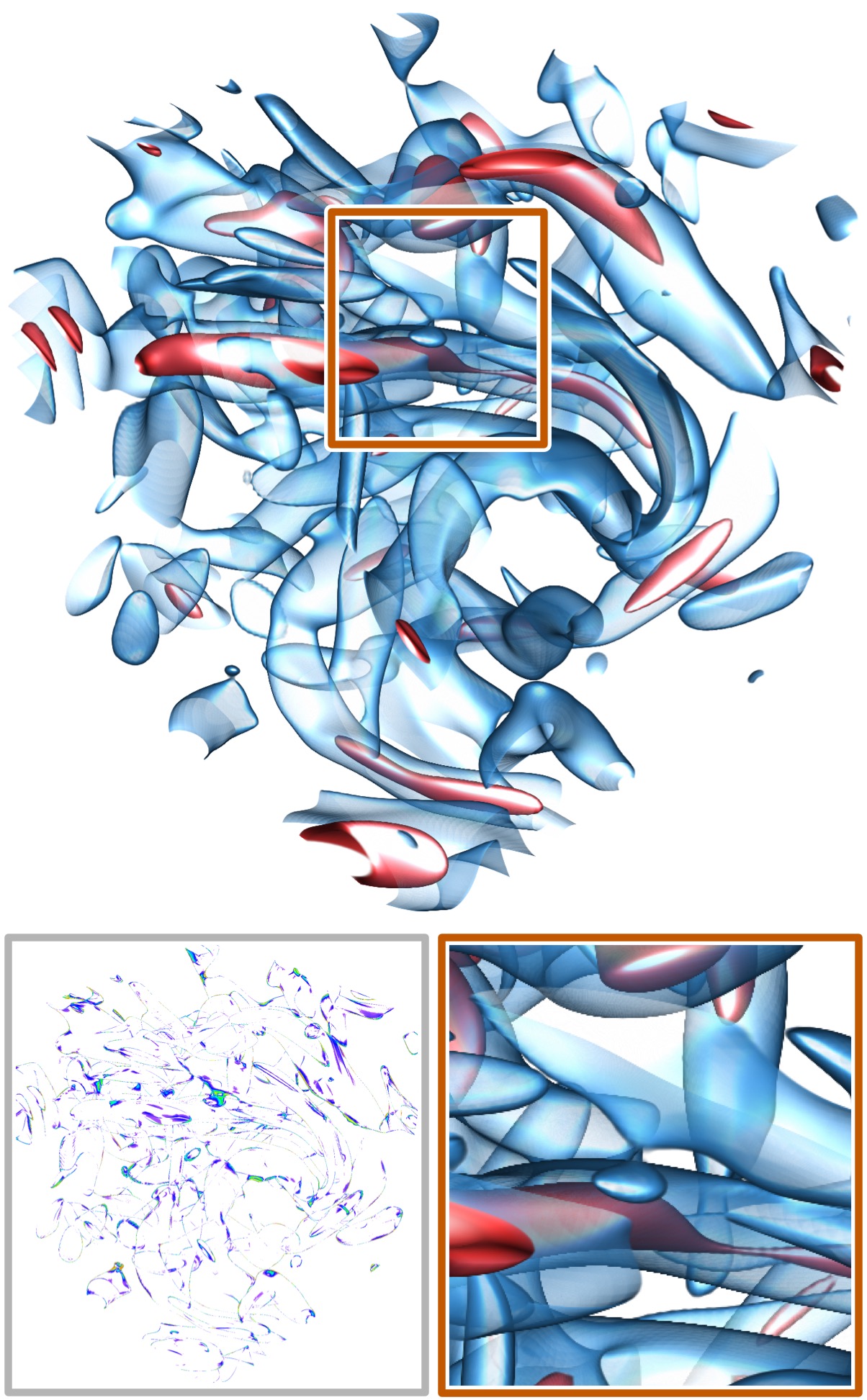}&
 \includegraphics[width=0.23\linewidth]{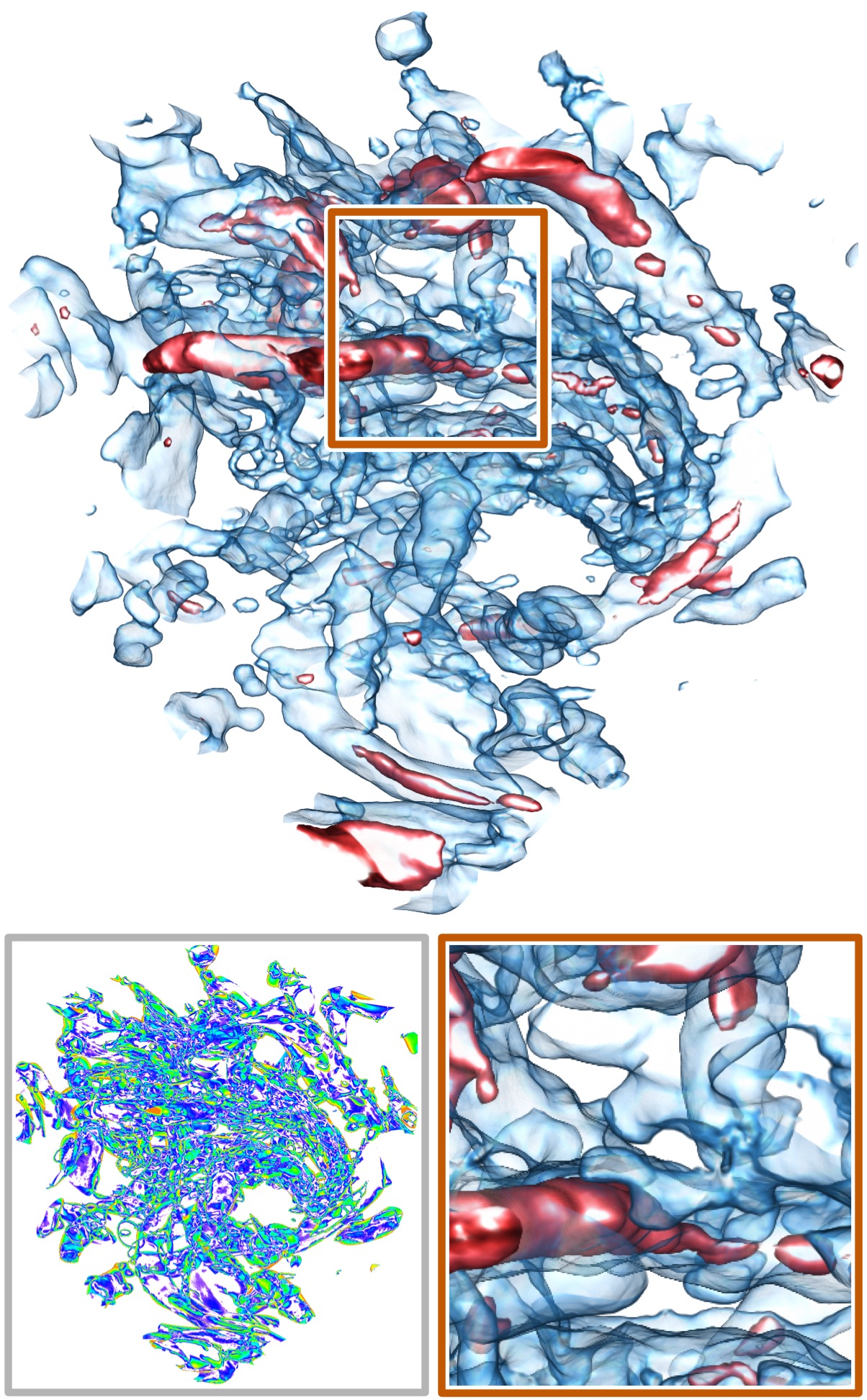}&
  \includegraphics[width=0.23\linewidth]{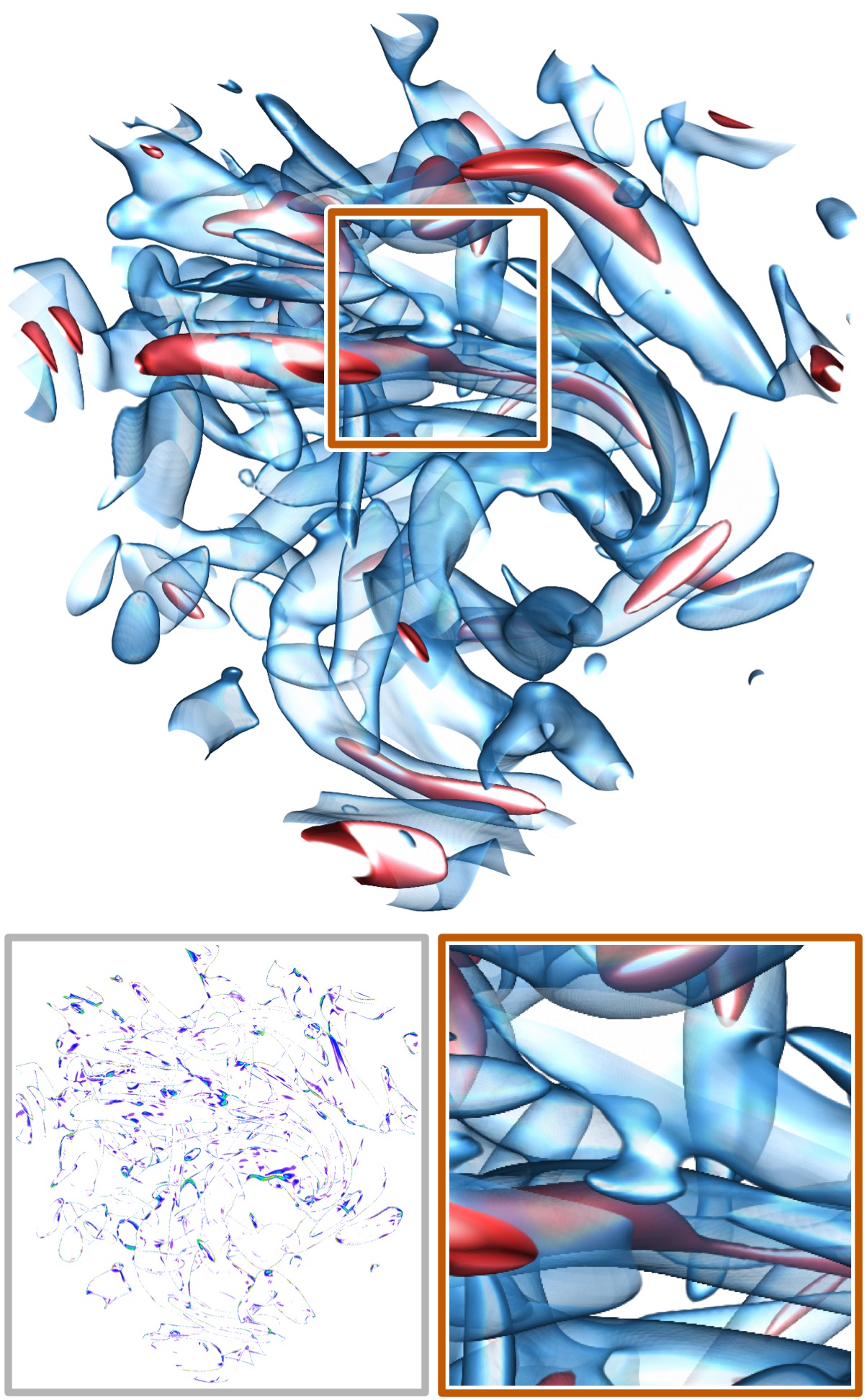}&
 \includegraphics[width=0.23\linewidth]{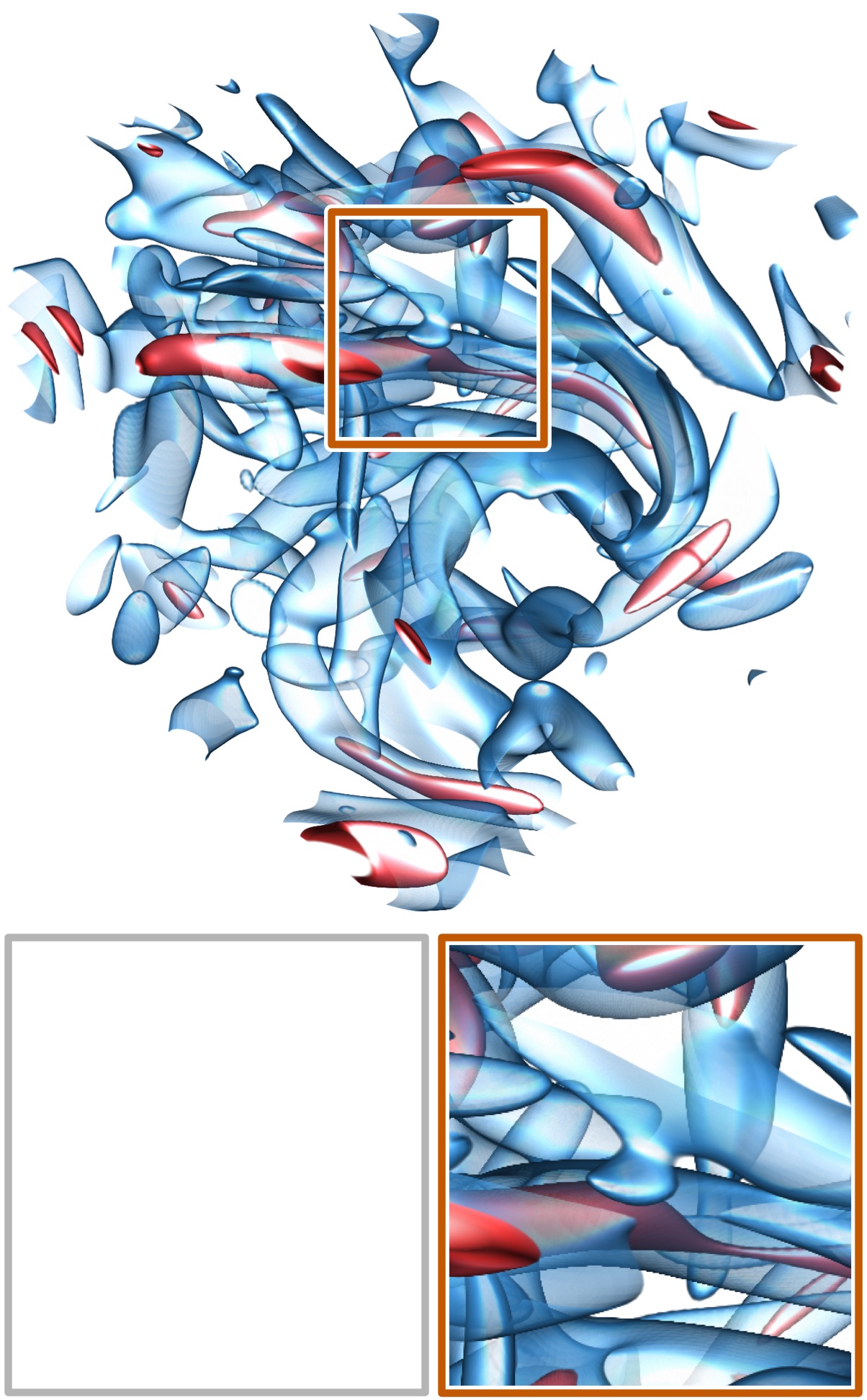} \\
\mbox{\footnotesize SIREN} & \mbox{\footnotesize p.t.\ SIREN} & \mbox{\footnotesize Meta-INR} &\mbox{\footnotesize GT}
\end{array}$
\end{center}
\vspace{-.25in} 
\caption{Comparing different methods on volume rendering results. Top to bottom: half-cylinder, ionization, Tangaroa, and vortex.} 
\label{fig:time-varying-vol}
\end{figure}

\begin{figure}[t]
 \begin{center}
 $\begin{array}{c@{\hspace{0.025in}}c@{\hspace{0.025in}}c@{\hspace{0.025in}}c}
\includegraphics[width=0.23\linewidth]{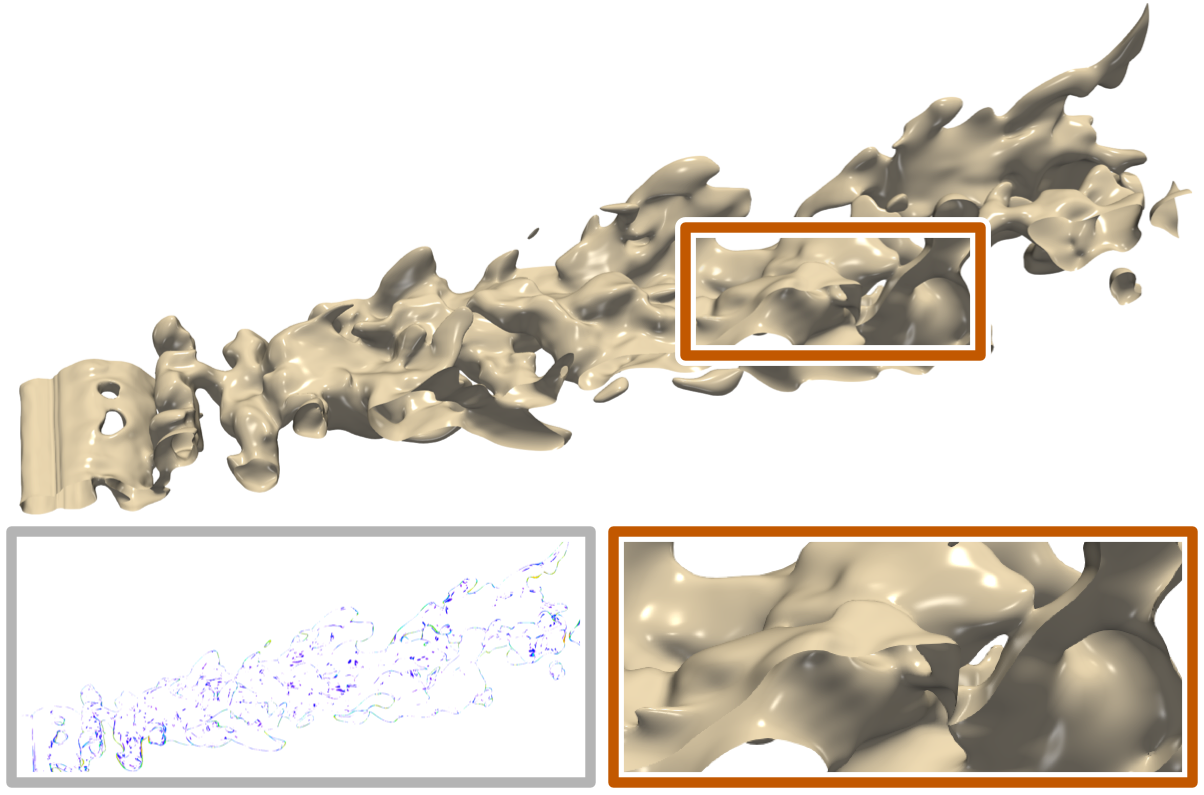}&
\includegraphics[width=0.23\linewidth]{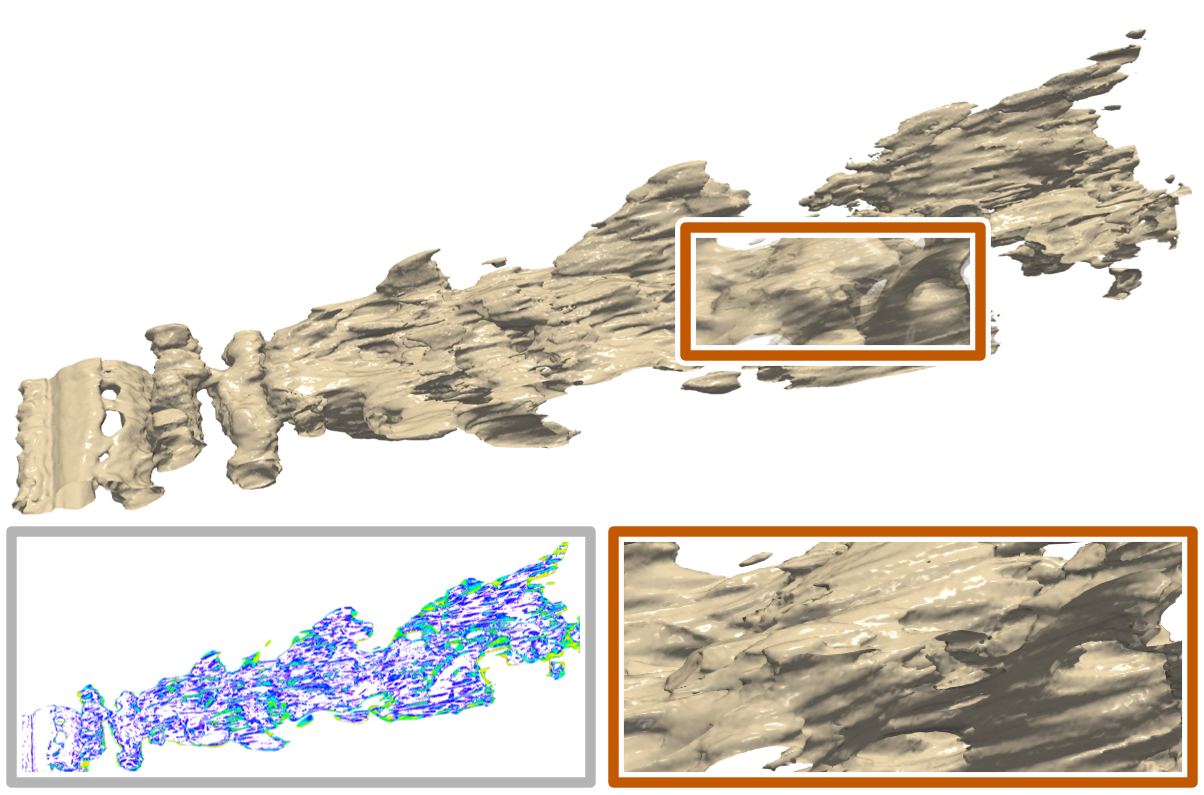}&
\includegraphics[width=0.23\linewidth]{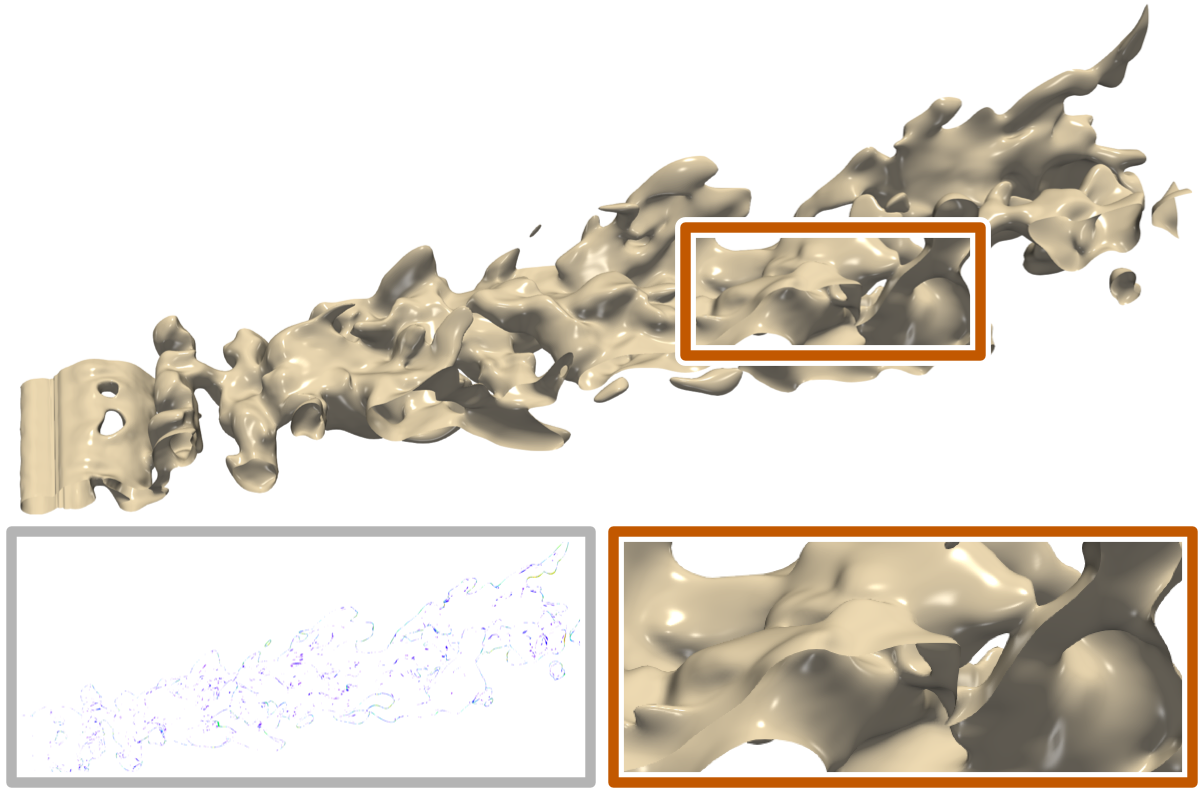}&
\includegraphics[width=0.23\linewidth]{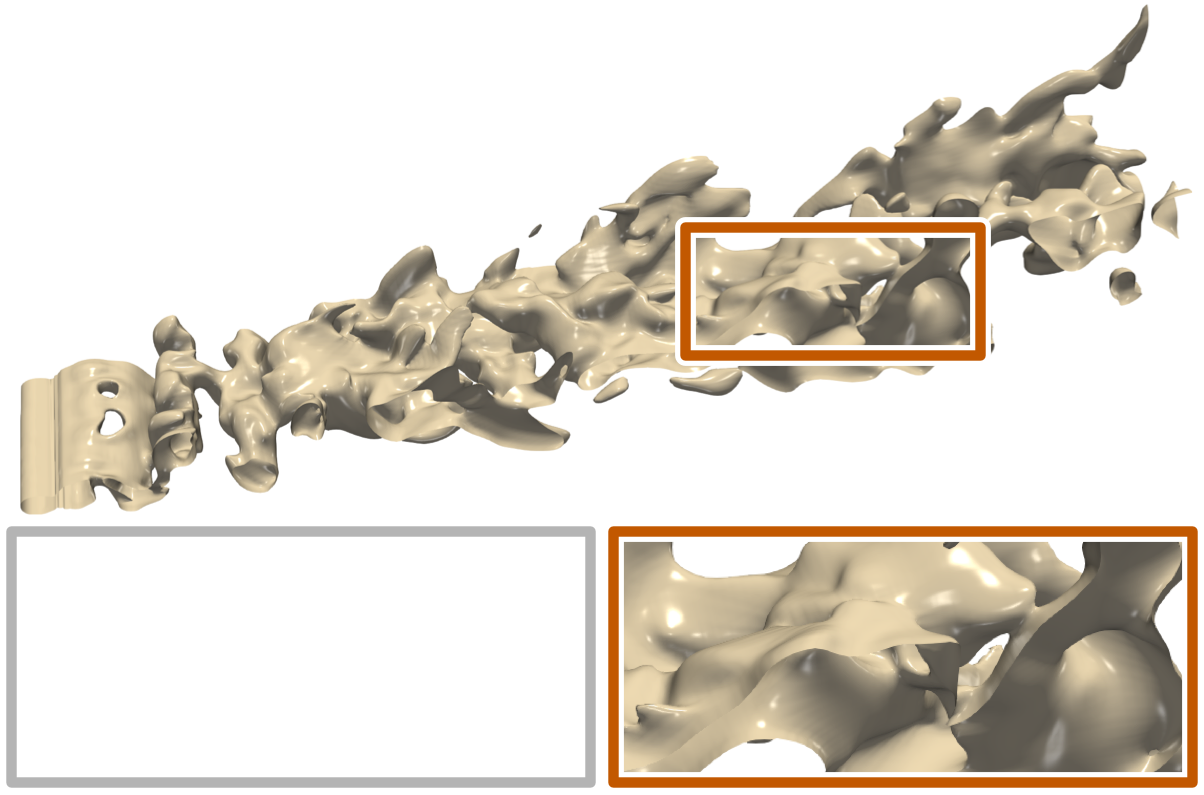} \\
\includegraphics[width=0.23\linewidth]{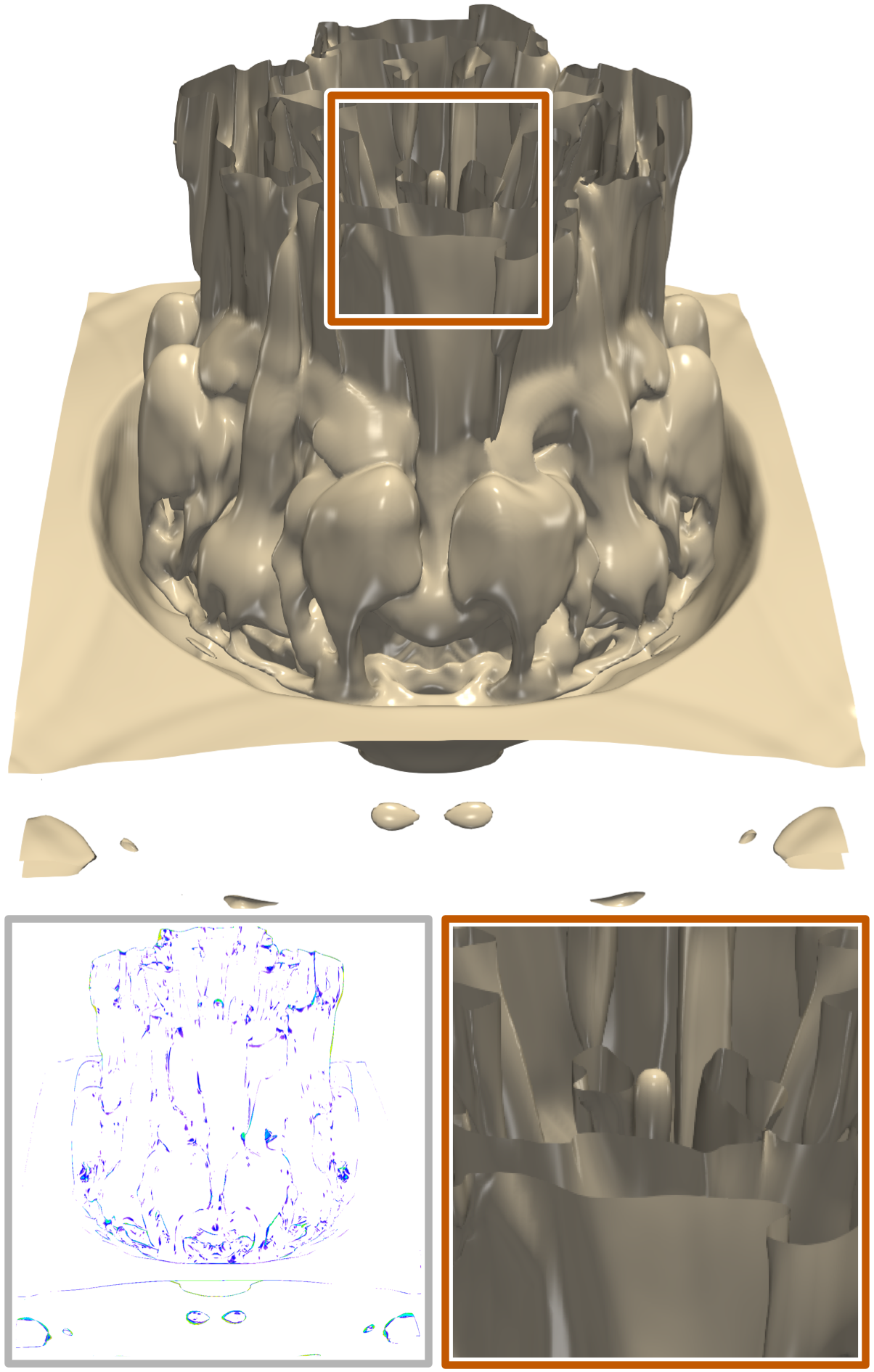}&
\includegraphics[width=0.23\linewidth]{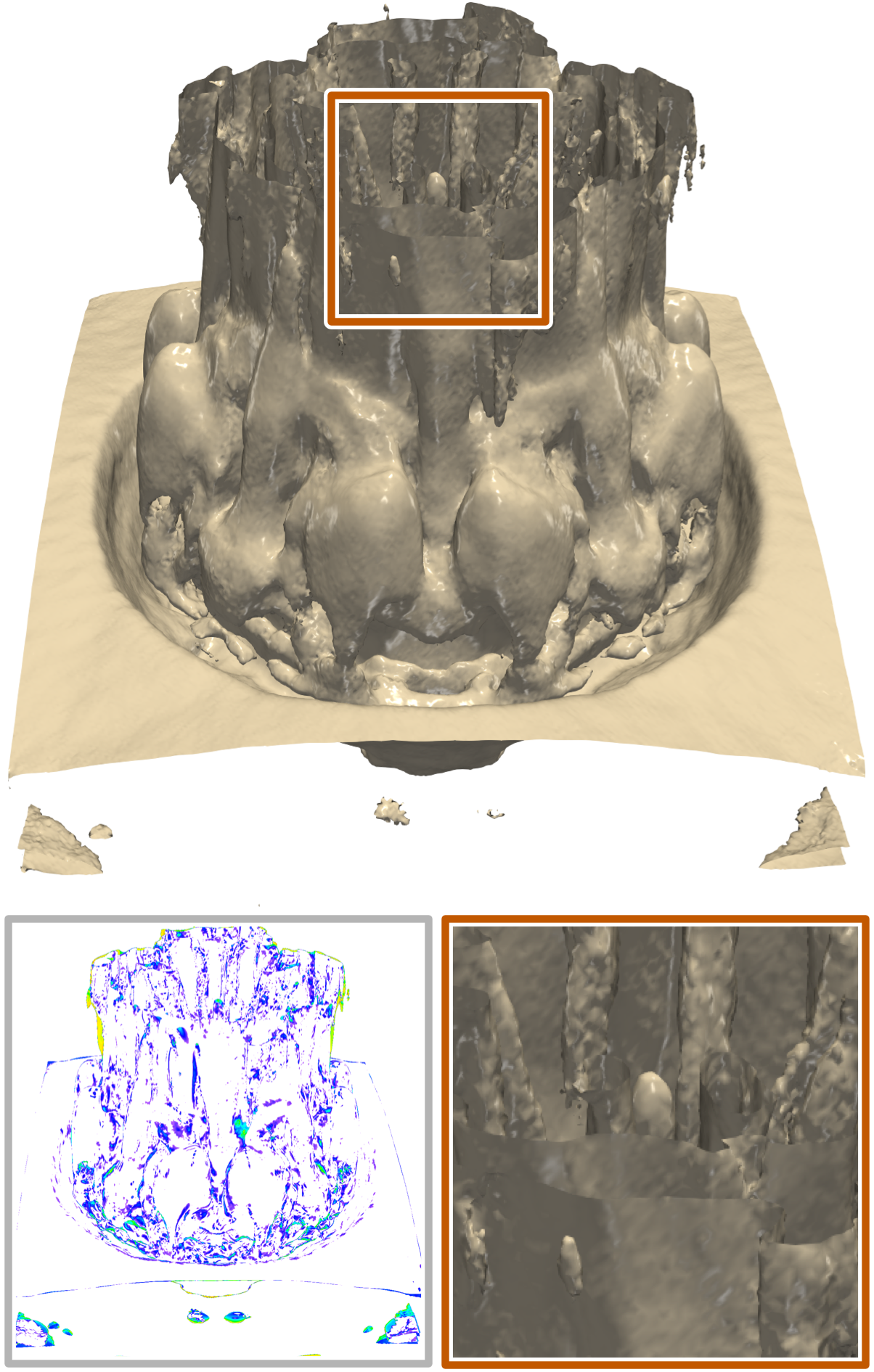}&
\includegraphics[width=0.23\linewidth]{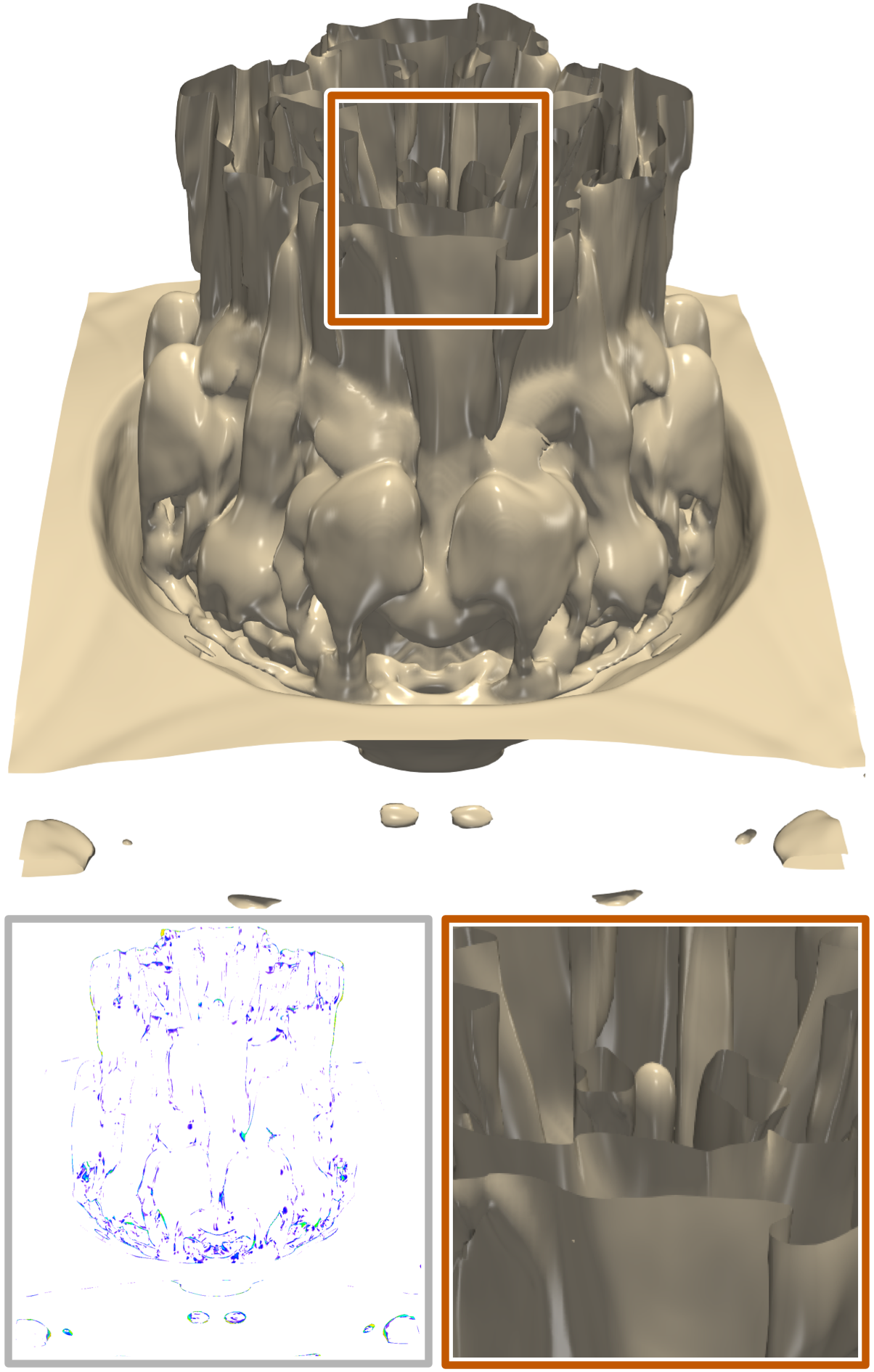}&
\includegraphics[width=0.23\linewidth]{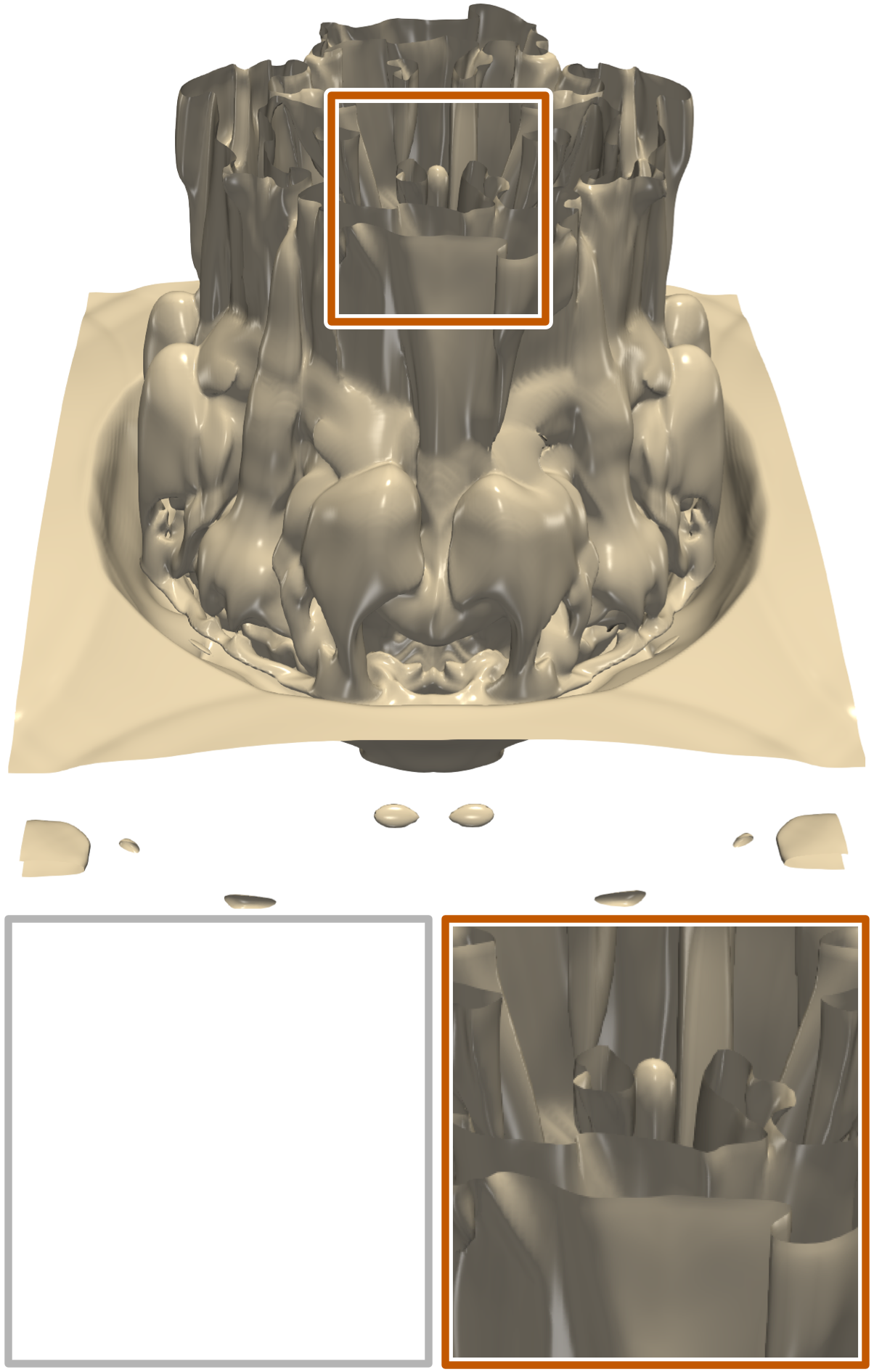} \\
\includegraphics[width=0.23\linewidth]{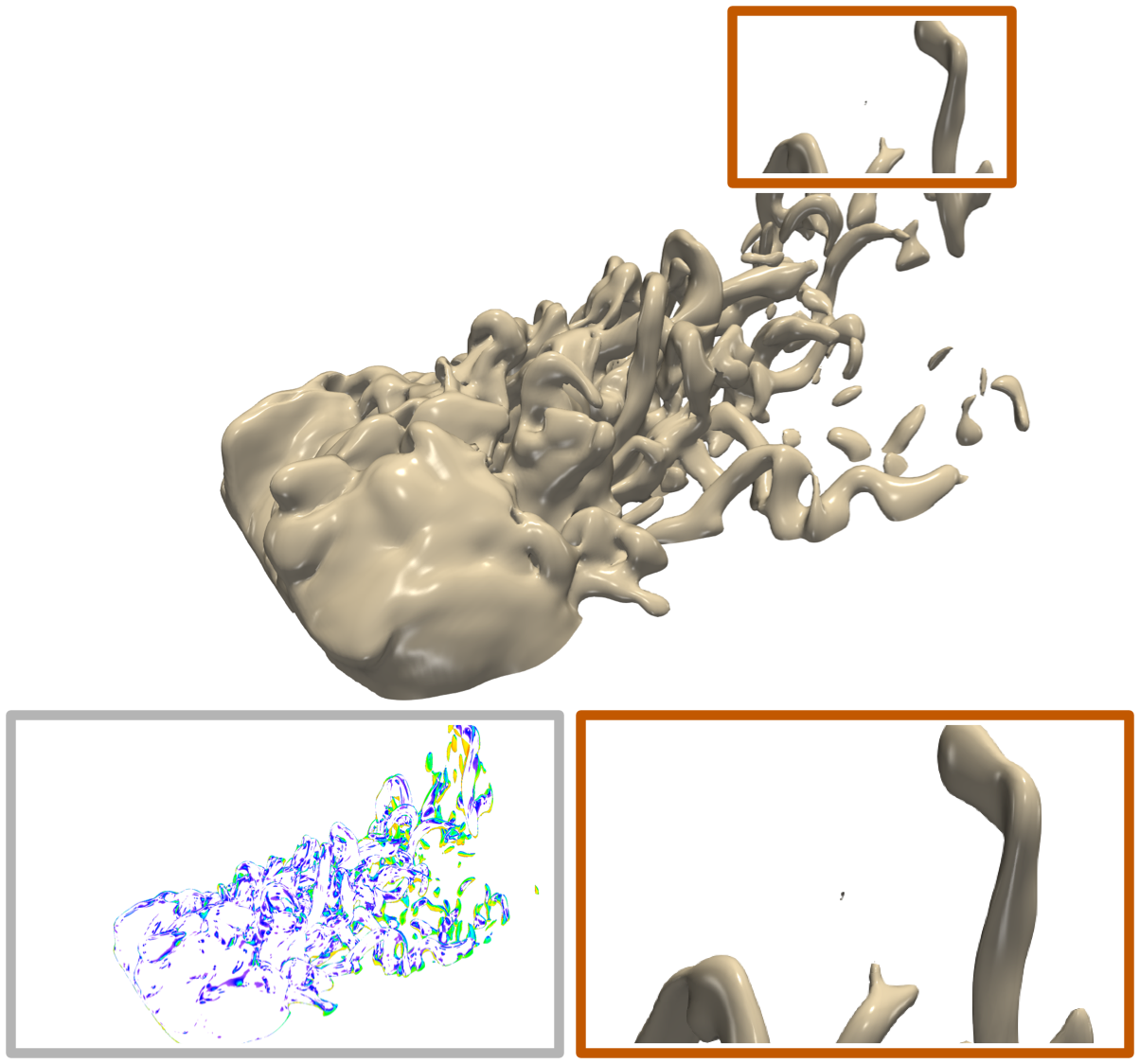}&
\includegraphics[width=0.23\linewidth]{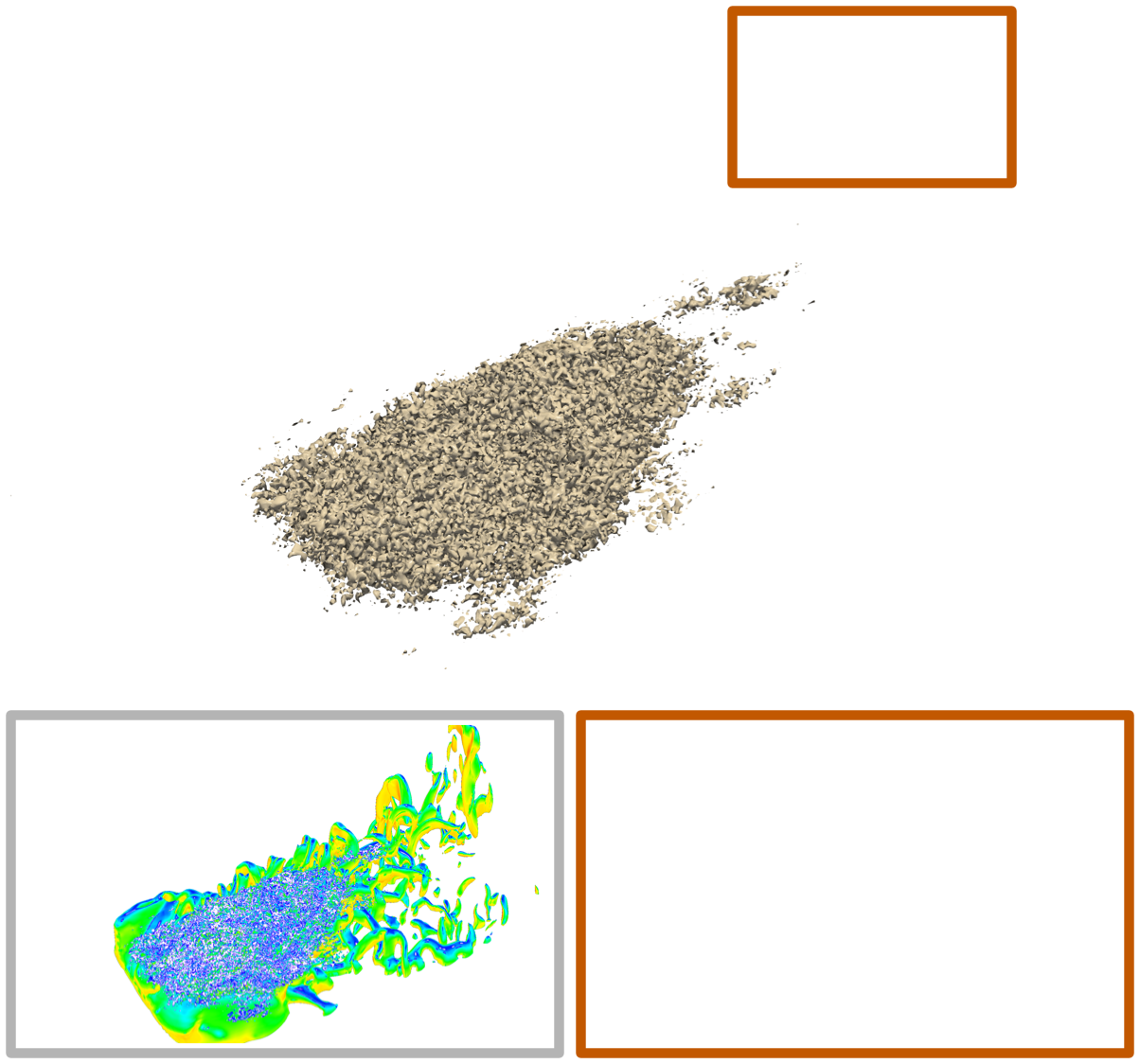}&
\includegraphics[width=0.23\linewidth]{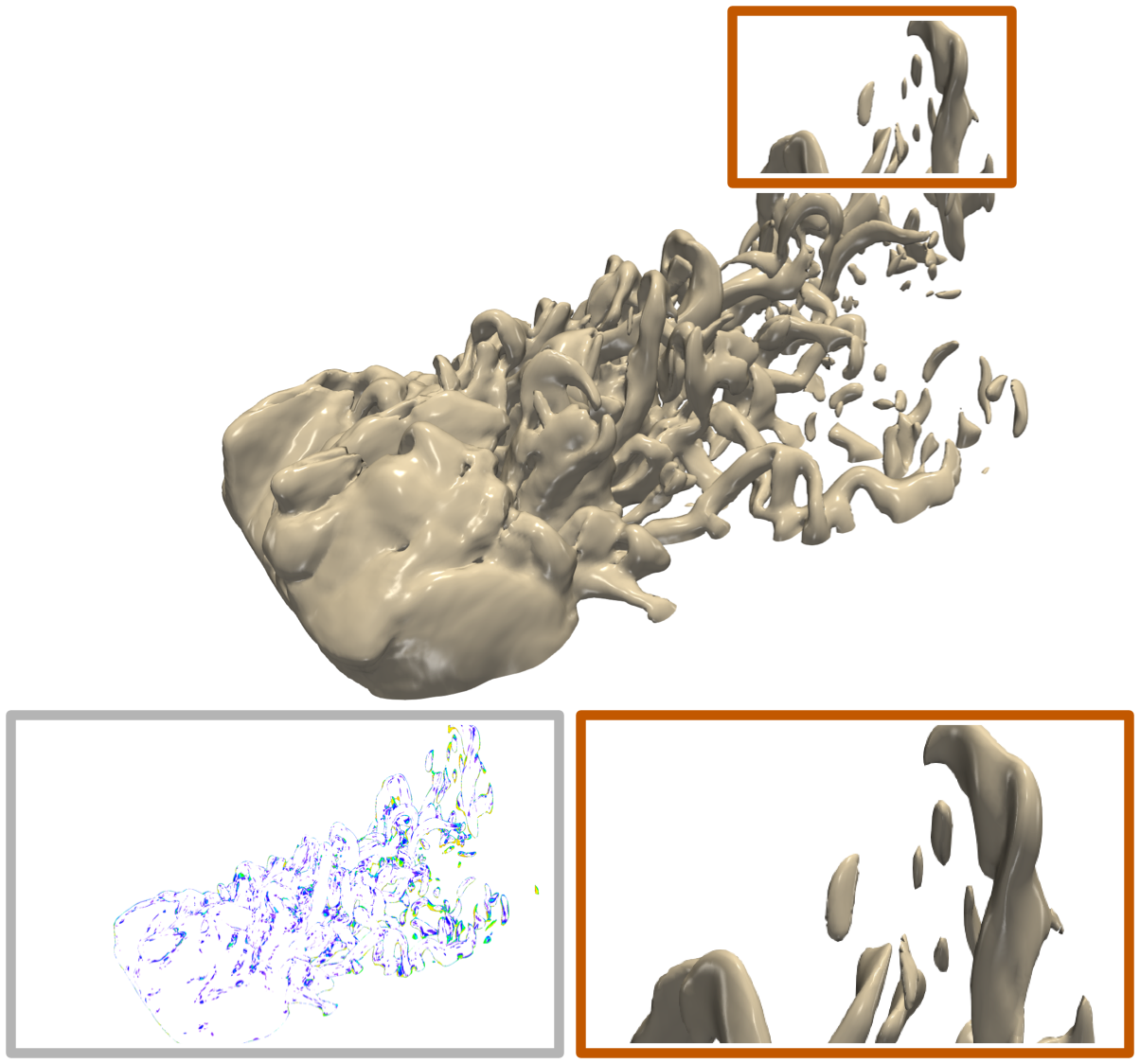}&
\includegraphics[width=0.23\linewidth]{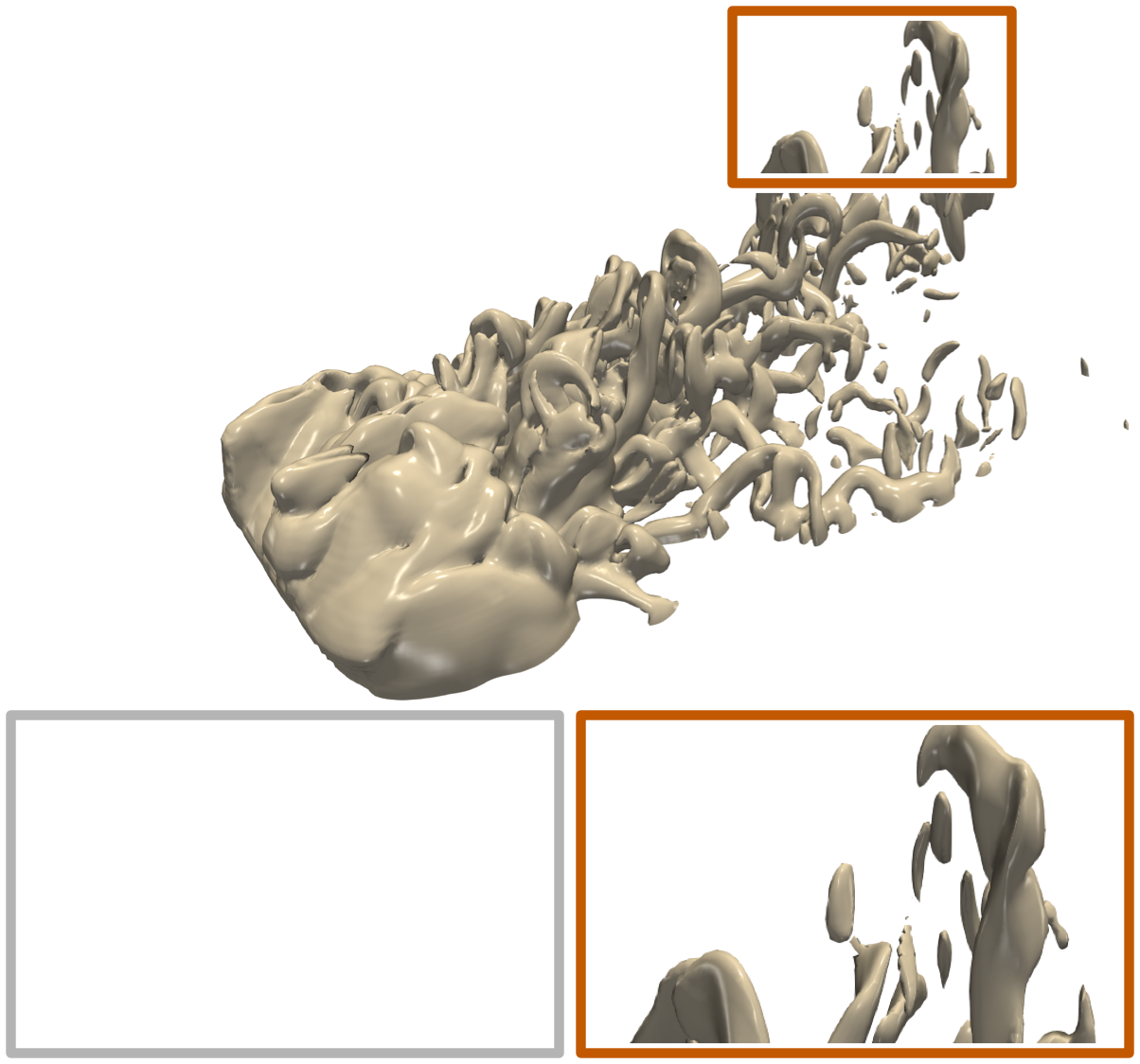} \\
\includegraphics[width=0.23\linewidth]{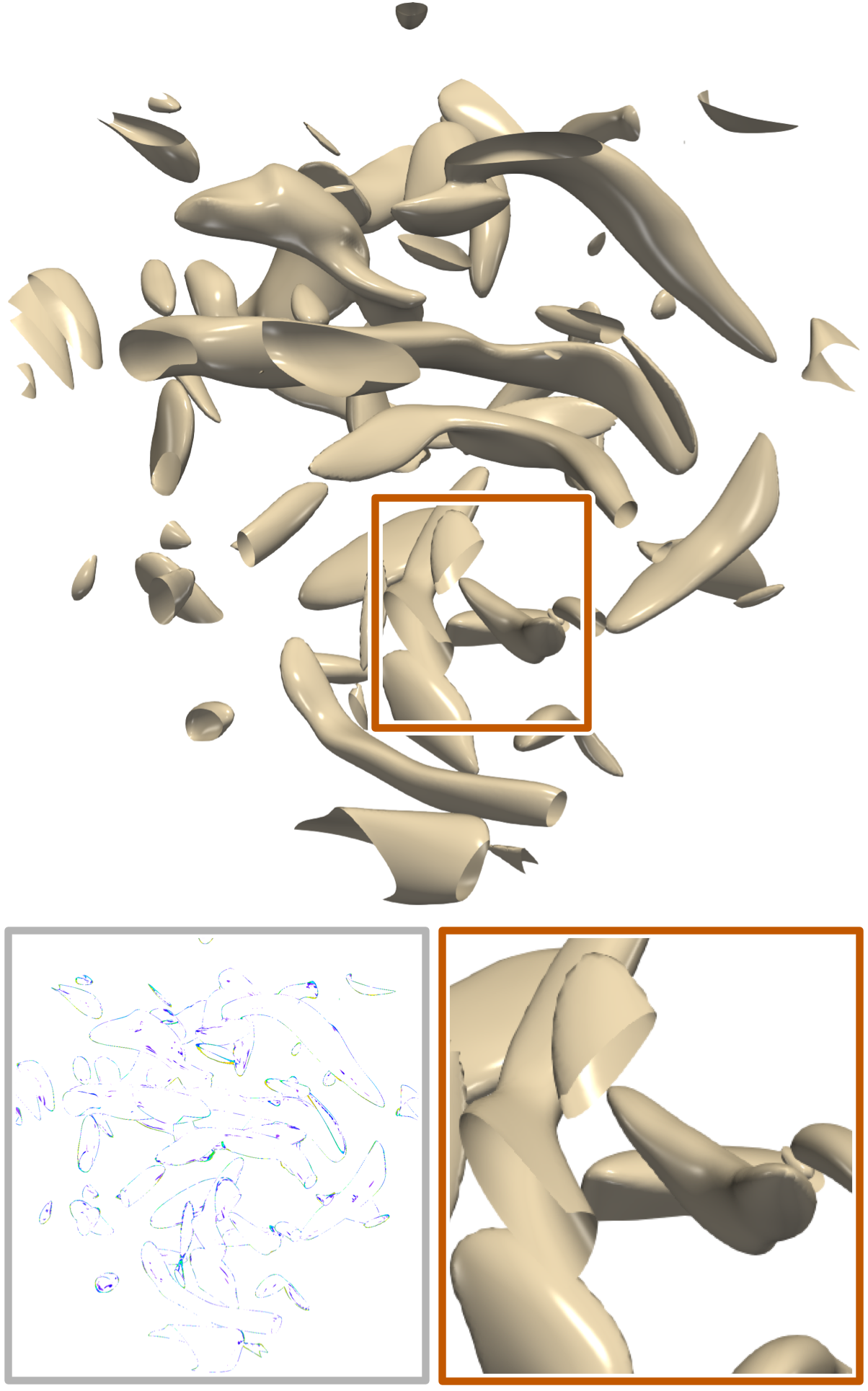}&
\includegraphics[width=0.23\linewidth]{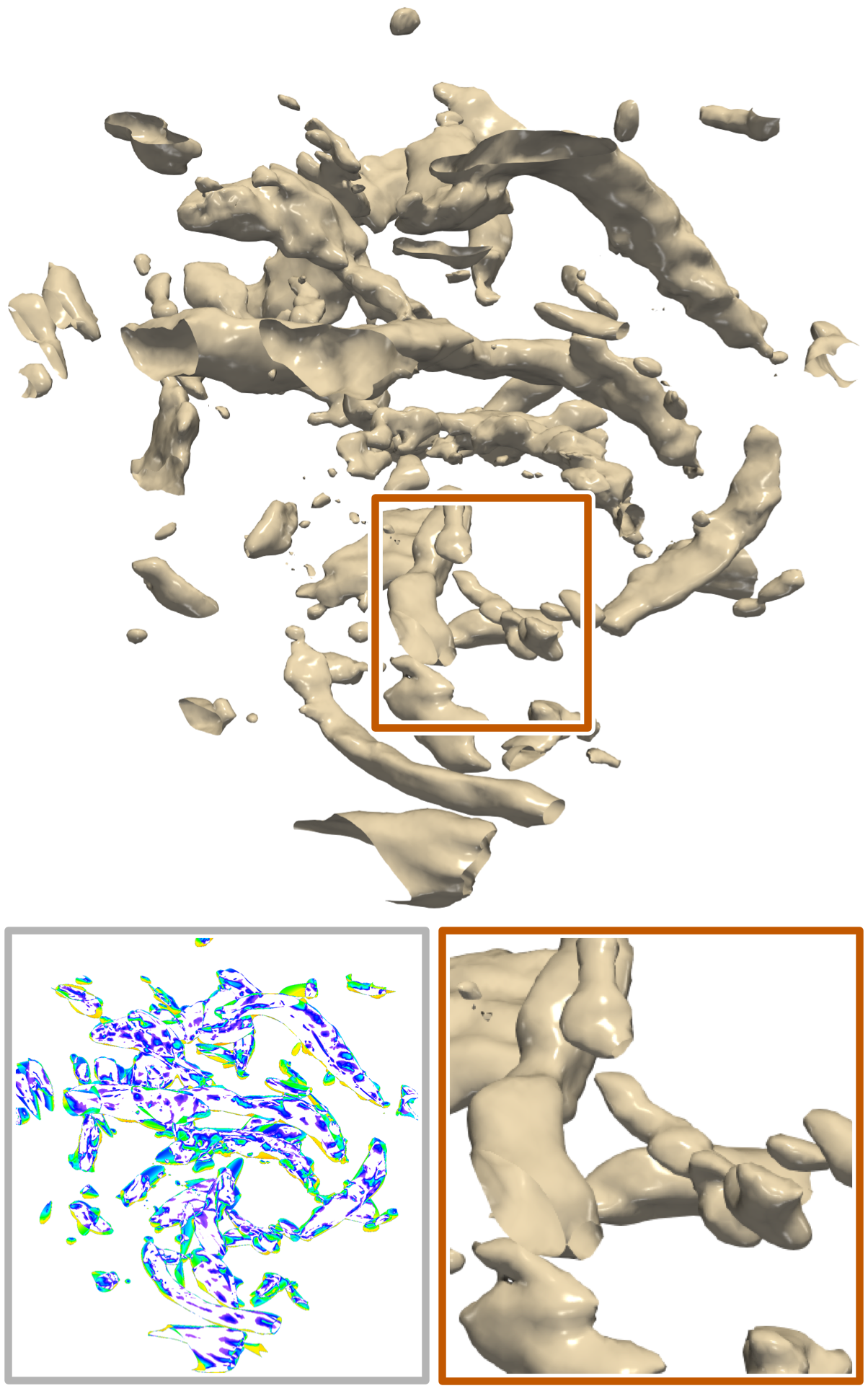}&
\includegraphics[width=0.23\linewidth]{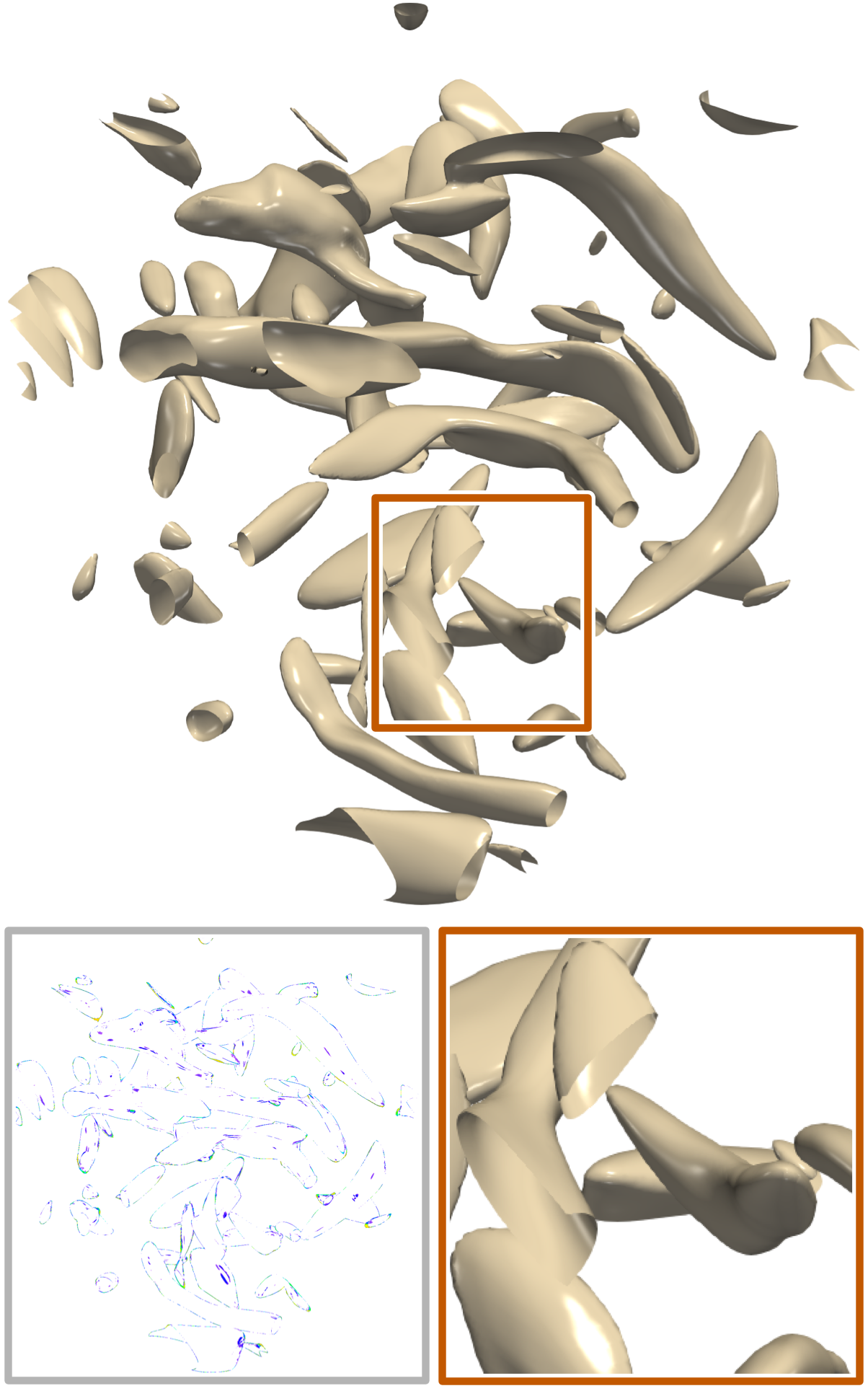}&
\includegraphics[width=0.23\linewidth]{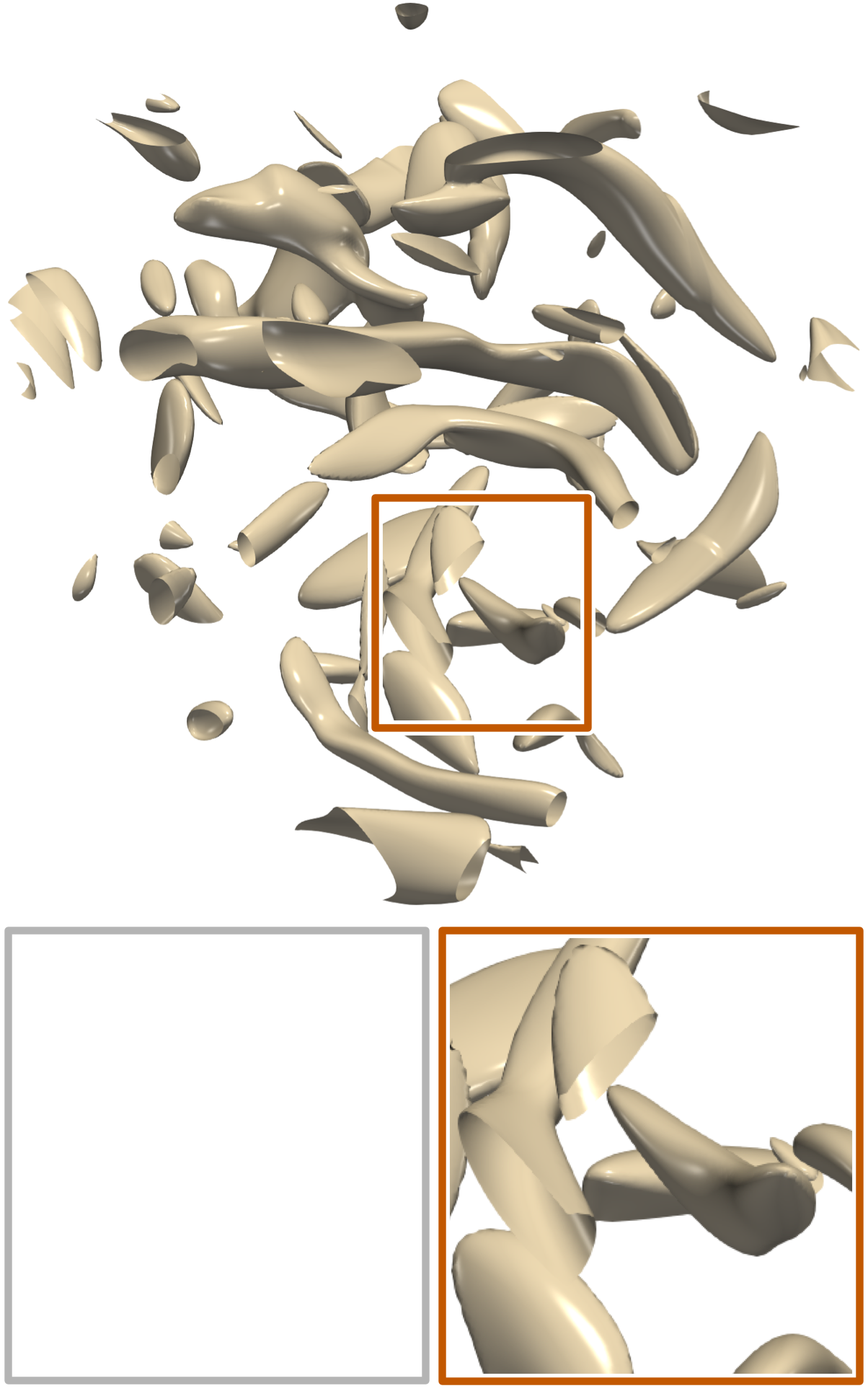} \\
\mbox{\footnotesize SIREN} & \mbox{\footnotesize p.t.\ SIREN} & \mbox{\footnotesize Meta-INR} &\mbox{\footnotesize GT}
\end{array}$
\end{center}
\vspace{-.25in} 
\caption{Comparing different methods on isosurface rendering results. Top to bottom: half-cylinder, ionization, Tangaroa, and vortex. The chosen isovalues are reported in Table~\ref{tab:baseline-metrics}.} 
\label{fig:time-varying-iso}
\end{figure}

\begin{figure*}[!ht]
\begin{center}
$\begin{array}{c@{\hspace{0.25in}}c@{\hspace{0.25in}}c}
\includegraphics[height=1.6in]{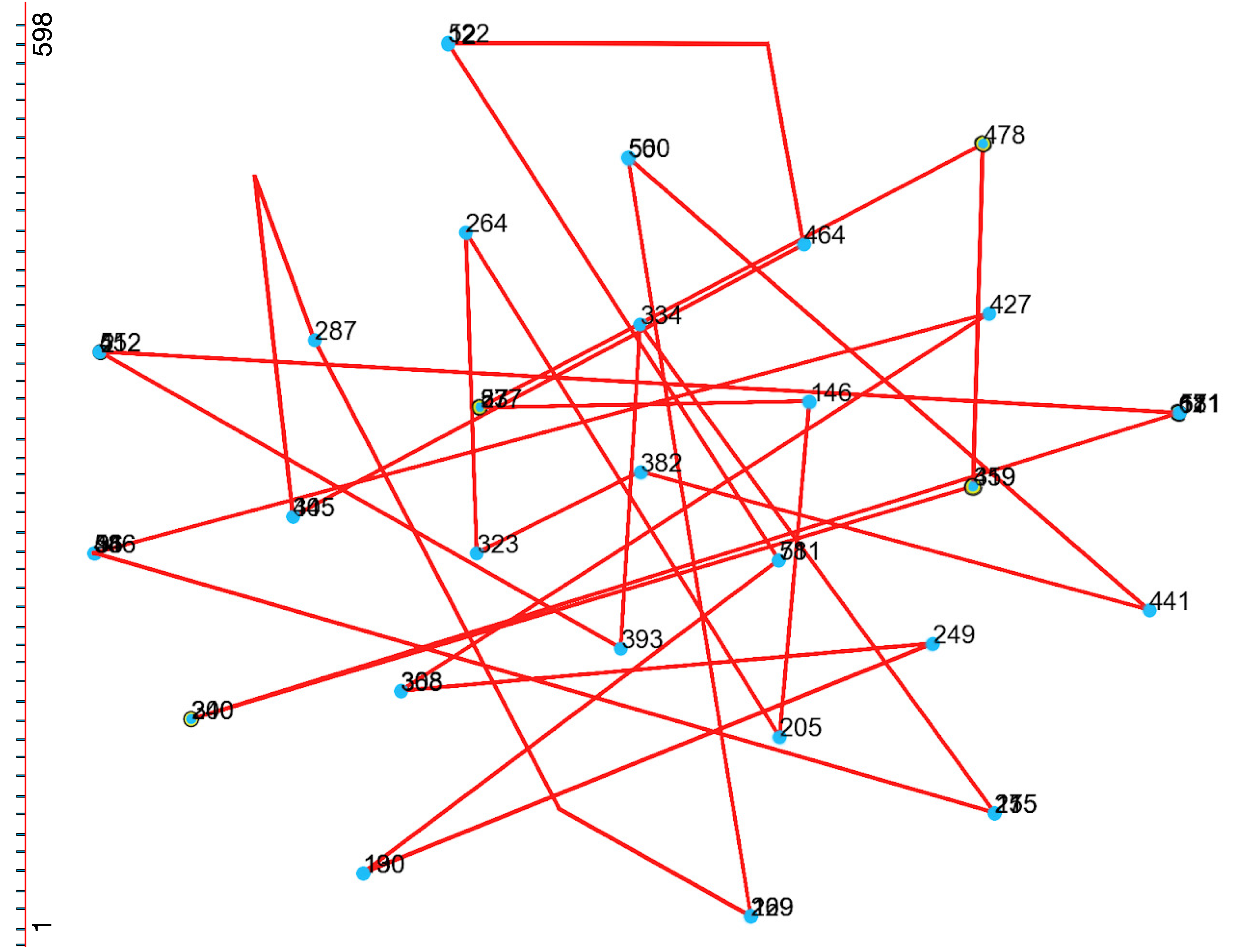} &
\includegraphics[height=1.6in]{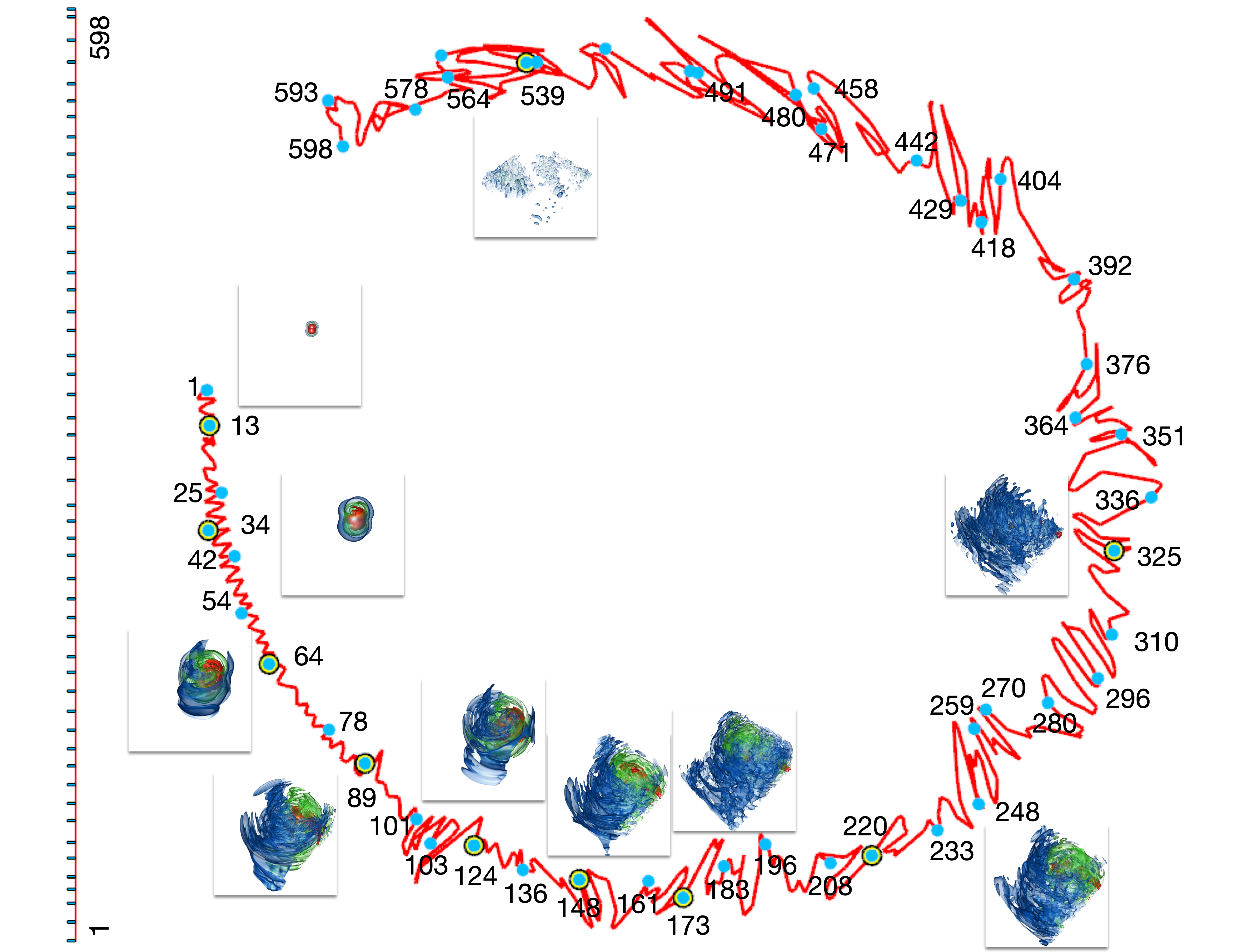} &
\includegraphics[height=1.6in]{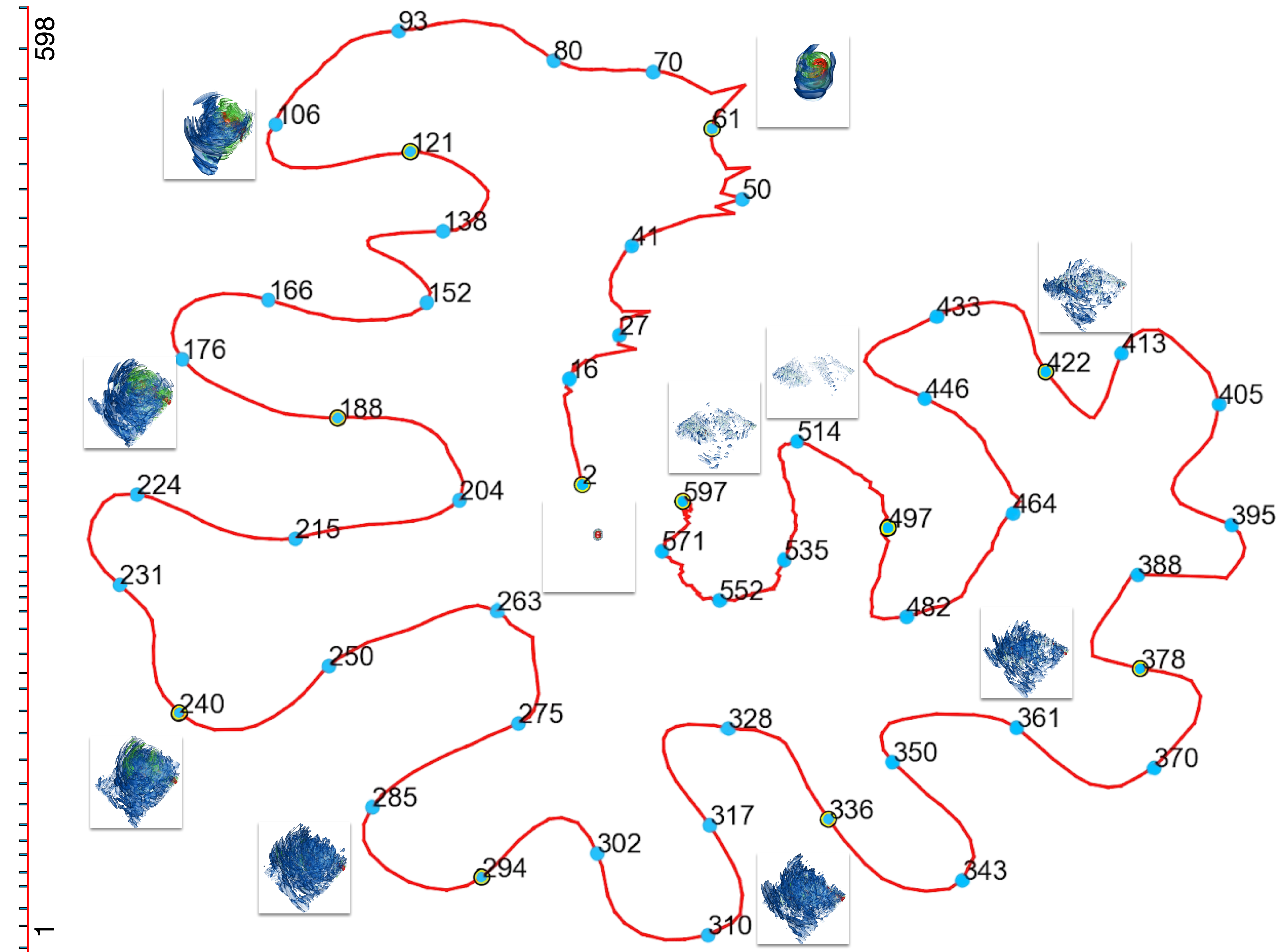} \\
\mbox{\footnotesize (a) SIREN~\cite{Sitzmann-SIREN-NeurIPS20}} & \mbox{\footnotesize (b) autoencoder~\cite{Porter-VISSP19}} &
\mbox{\footnotesize (c) Meta-INR} 
\end{array}$
\end{center}
\vspace{-.25in}
\caption{t-SNE projections using parameters from different methods trained on the time-varying earthquake dataset, where selected timesteps are marked on the timeline on the left. SIREN's model parameters are not interpretable and fail to establish a correlation between timesteps. Volume rendering images are shown adjacent to selected timesteps for autoencoder and Meta-INR. 
}
\label{fig:timestep-selection}
\end{figure*}

\begin{figure*}[!htb]
\centering
\begin{center}
$\begin{array}{c@{\hspace{0.25in}}c@{\hspace{0.25in}}c@{\hspace{0.25in}}c}
\includegraphics[width=0.215\linewidth]{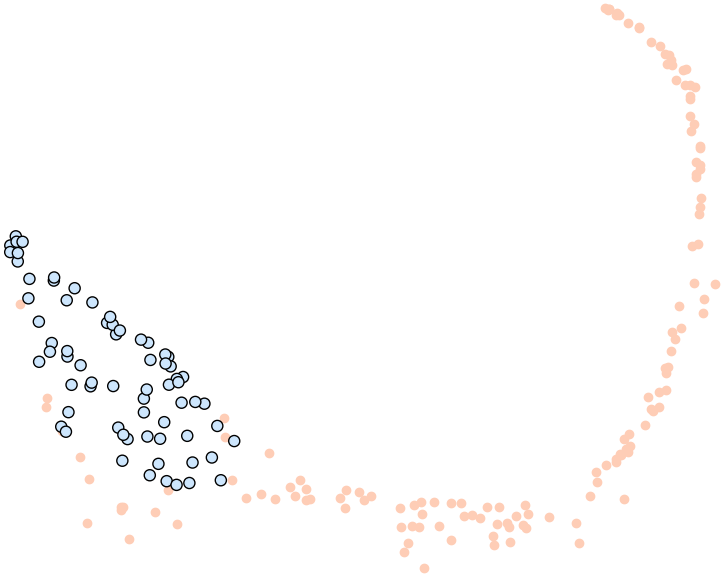} &
\includegraphics[width=0.215\linewidth]{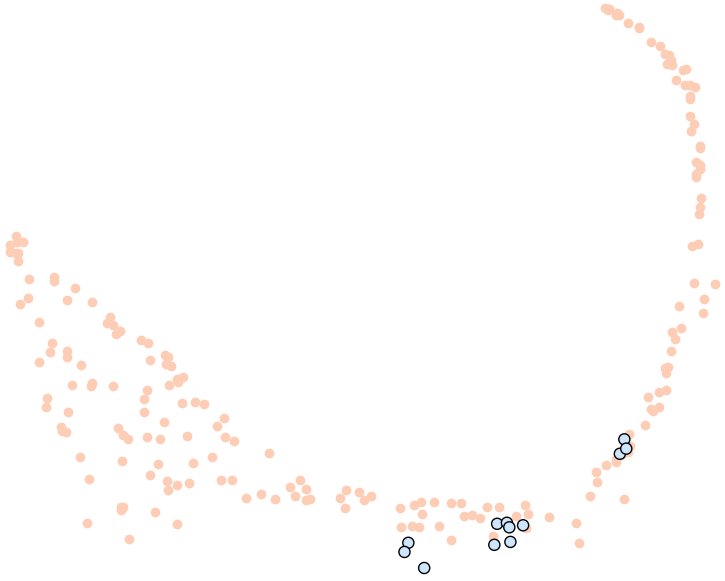} &
\includegraphics[width=0.215\linewidth]{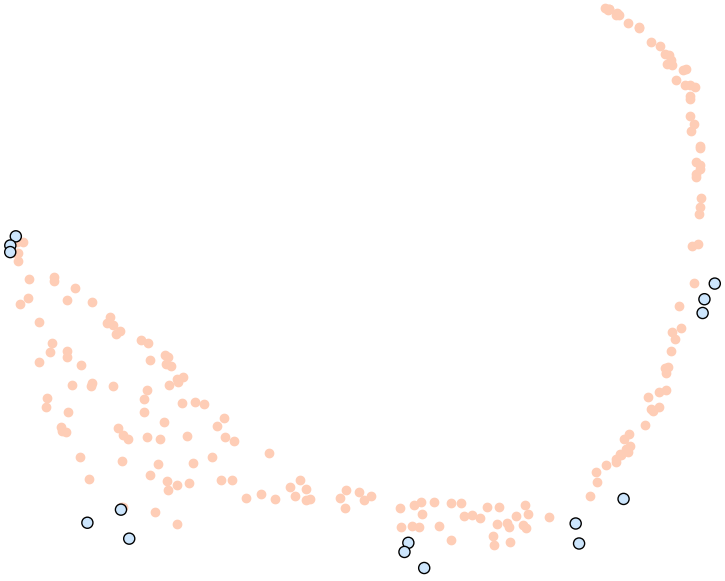} &
\includegraphics[width=0.215\linewidth]{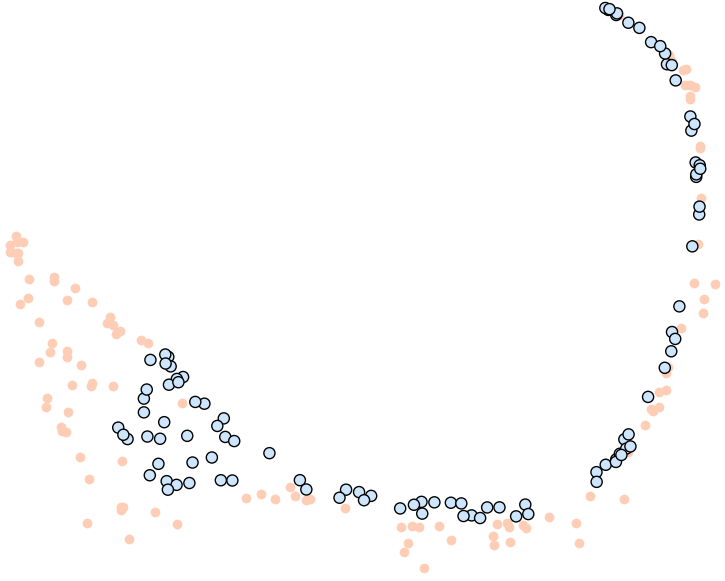} \\
\mbox{\footnotesize (a) $h_0=550,000$} & \mbox{\footnotesize (b) $h_0=700,000$} &
\mbox{\footnotesize (c) $OmM_0=120,000$} & \mbox{\footnotesize (d) $OmM_0=155,000$} 
\end{array}$
\end{center}
\vspace{-.25in}
\caption{t-SNE projections of Meta-INR models trained on the ensemble Nyx dataset where each point represents a volume. Volumes with $h_0$ or $OmM_0$ matching the specified values are highlighted in light blue.}
\label{fig:multi-variate}
\end{figure*}

\vspace{-0.1in}
\section{Results and discussion}

\subsection{Datasets, Training, Baselines, and Metrics}

{\bf Datasets and network training.}
Table~\ref{tab:datasets} lists the datasets used to evaluate Meta-INR. 
In particular, we utilize the time-varying datasets, including the half-cylinder, ionization, Tangaroa, and vortex datasets, to evaluate the reconstruction accuracy of the Meta-INR compared with other baseline methods.
The time-varying earthquake dataset is leveraged to showcase the interpretability of the parameters of adapted INRs under t-SNE projection.
We demonstrate the ability of Meta-INR on simulation parameter analysis using the ensemble Nyx dataset with three simulation parameters: $h_0$, $OmM_0$, and $OmB_0$.
Meta-INR uses a seven-layer SIREN model~\cite{Sitzmann-SIREN-NeurIPS20} as its network backbone, with a hidden layer dimension of 256.
It is trained across all experiments with 500 outer steps and sets the number of inner-loop steps $K=16$.
We set both inner and outer loop learning rates $\alpha$ and $\beta$ to 0.0001 during meta-pretraining and 0.00001 during volume-specific finetuning. 
Both stages optimize the parameters with a batch size of 50,000 coordinate-value pairs for each iteration.

{\bf Baselines.} We compare Meta-INR with two baseline strategies:
\begin{myitemize}
\vspace{-0.05in}
    \item SIREN~\cite{Sitzmann-SIREN-NeurIPS20} is the backbone network architecture of Meta-INR. Here, SIREN as a baseline method means training each INR network independently for each volume from scratch.
    \item Pretrained SIREN is a vanilla pretraining baseline method that optimizes the initial parameters on the same subsampled dataset without leveraging meta-learning techniques (i.e., no inner-loop updating). After pretraining, we finetune the learned initial parameters using the same number of adaptation steps as the Meta-INR for fair comparisons.

\vspace{-0.05in}
\end{myitemize}

{\bf Evaluation metrics.}
We use three metrics to evaluate the reconstruction accuracy of Meta-INR and baseline methods. 
We utilize \textit{peak signal-to-noise ratio} (PSNR) to measure the volume reconstruction accuracy, and \textit{learned perceptual image patch similarity} (LPIPS) to evaluate rendered image qualities for volumes reconstructed by different methods. \textit{Chamfer distance} (CD) is used to assess similarity at the surface level by calculating the average distance of isosurfaces. 

\vspace{-0.05in}
\subsection{Time-Varying Data Representation}
\label{subsec:tvdr}

Time-varying data representation directly evaluates the effectiveness of Meta-INR in speeding up model training and improving reconstruction fidelity. We consider all methods on four datasets with various dimensions, as seen in Table~\ref{tab:datasets}. 

{\bf Quantitative comparison.}
In Table~\ref{fig:time-varying-iso}, we report quantitative results of all different methods. 
Meta-INR performs the best across the three quality metrics for all datasets. 
Although SIREN achieves decent quality, sometimes comparable to Meta-INR, the need to retrain a model from scratch for each volume implies that a significant encoding time is required for the model to converge. 
For Meta-INR, despite the model taking the majority of total time for meta-pretraining, only a few adaptation steps are needed for encoding, saving considerable time compared to SIREN.
In particular, the encoding time of adapted INRs is 5.87$\times$ faster on average across all datasets than SIREN, which trains each INR from scratch. 
Pretrained SIREN performs the worst in terms of the three quality metrics as it fails to capture the intrinsic patterns from partial observation of volumetric datasets and struggles to converge efficiently during encoding.
Through diverse evaluation across multiple datasets varying in size and complexity, we found Meta-INR to be highly adaptable. Meta-INR's generalizability stems from its meta-learning framework, which uses inner loop adaptation during meta-pretraining that mimics the finetuning process, forcing parameters into a region where a small number of gradient steps can minimize task-specific loss. This design ensures the prior captures shared structural patterns while remaining sensitive to volume-specific variations. 

{\bf Qualitative comparison.} 
In Figures~\ref{fig:time-varying-vol}~and~\ref{fig:time-varying-iso}, we compare the volume and isosurface rendering results for different methods. 
A pixel-wise difference image is also provided on the bottom left to show the differences between each method and the GT.
Pretrained SIREN performs the worst among all three methods, showing significant deviations from the GT. In many cases, it demonstrates block-like artifacts. 
Both SIREN and Meta-INR demonstrate excellent visual fidelity for the reconstructed volumes. Although they perform similar results in simple datasets such as vortex, Meta-INR can achieve significantly higher accuracy for complex ones with many details, such as Tangaroa. This is because such details are already captured in the meta-model, which serves as the prior for quick adaptation. Then, during volume-specific finetuning, only slight parameter adaptation is needed.

\vspace{-0.05in}
\subsection{Representative Timestep Selection}

We showcase the interpretability of Meta-INR by analyzing its ability on the representative timestep selection task of the earthquake dataset. 
Porter et al.\ \cite{Porter-VISSP19} demonstrated the effectiveness of deep learning techniques for selecting representative timesteps.
In our scenario, an INR network trained on one specific volume can also be considered an alternative representation of that volume.
However, as shown in Figure~\ref{fig:timestep-selection}, when we use t-SNE~\cite{t-SNE} to project the learned parameters of each SIREN at each timestep to a 2D space and connect the points in the order of timesteps, the resulting projection view is meaningless and offers no interpretability. 
This is because of significantly increased noise and randomness in training, which leads to models finding drastically different local minima. In contrast, Meta-INR's volume-specific finetuning starts with a common prior, eliminating most noise and focusing on each volume's differences.
In particular, the connected points of SIREN across different timesteps lack continuity, resulting in numerous sharp turnings. 
As SIREN encodes volumes at each timestep independently from scratch, their parameter representations fail to capture a smooth transition of the dataset along the time dimension.
Unlike SIREN, the connected points are meaningful and smooth when leveraging t-SNE to project the parameters of adapted INRs finetuned on each timestep.
We select representative timesteps following~\cite{Porter-VISSP19} and observe reasonable results.
Moreover, we can see that the connected points of Meta-INR are smoother than the results obtained by~\cite{Porter-VISSP19} using an autoencoder.
This difference arises from the need to downsample the data before training the autoencoder. 
Unlike Meta-INR, the network architecture of the autoencoder is constrained by volume dimensions and cannot directly process high-resolution volumetric data.
Therefore, the connected points for the projections of the autoencoder are less smooth due to the noise introduced from downsampling.

\vspace{-0.05in}
\subsection{Simulation Parameter Analysis}
\label{subsec:spa}

When analyzing ensemble datasets, it is often insightful to see how changes in each parameter affect the simulation. 
Meta-INR can aid in this process by effectively visualizing the relative differences between each volume via t-SNE projection. We apply Meta-INR to the Nyx dataset with three simulation parameters, $h_0$, $OmM_0$, and $OmB_0$. Similar to the time-varying datasets, we perform meta-pretraining on the subsampled Nyx dataset, which is downsampled along the spatial and ensemble dimensions, and then conduct volume-specific finetuning to fit all volumes corresponding to different simulation parameters. In Figure~\ref{fig:multi-variate}, we visualize the pattern in model parameters using t-SNE and mark t-SNE projections sharing the same parameter values. 
In (a) and (b), we highlight all projections with $h_0$ equal to 550,000 and 700,000, respectively. We can see that $h_0=550,000$ corresponds to projections centralized on the left side, and the projections of $h_0=700,000$ are centralized at the bottom of the plot.
Similarly, in (c) and (d), which mark projections with $OmM_0$ equal  120,000 and 155,000, respectively, we observe that $OmM_0=120,000$ corresponds to projections located at the outer region of the plot. On the other hand, $OmM_0=155,000$ corresponds to projections close to the inner region.
These results show that the parameters of adapted INRs assimilate information about the simulation parameters during the encoding process.

\vspace{-0.075in}
\section{Conclusions and future work}
\vspace{-0.025in}

We have presented Meta-INR, a pretraining method designed to optimize a meta-model that can adapt to unseen volume data efficiently.
The generalizability of the meta-model allows for fast convergence during volume-specific finetuning while retaining the model interpretability within its parameters. 
The evaluation of various volumetric data representation tasks demonstrates better quantitative and qualitative performance of the meta-pretraining than other training strategies.  

For future work, we would like to explore continual learning to improve the capabilities of the meta-model for handling more complex time-varying volume data with significantly larger timesteps and variations.
Moreover, we want to investigate meta-pretraining on grid-based INRs such as fV-SRN~\cite{Weiss-CGF22} or APMGSRN~\cite{Wurster-TVCG24}.
Finally, we plan to incorporate transfer learning techniques to learn potentially more challenging variations across variables for multivariate datasets, allowing a meta-model trained on one variable sequence to be effectively used in finetuning another.

\vspace{-0.05in}
\acknowledgments{
This research was supported in part by the U.S.\ National Science Foundation through grants IIS-1955395, IIS-2101696, OAC-2104158, and IIS-2401144, and the U.S.\ Department of Energy through grant DE-SC0023145. The authors would like to thank the anonymous reviewers for their insightful comments.
}

\vspace{-0.05in}
\bibliographystyle{abbrv-doi-hyperref}
\bibliography{template}

\begin{thebibliography}{10}

\bibitem{Almgren-AJ13}
A.~S. Almgren, J.~B. Bell, M.~J. Lijewski, Z.~Luki{\'c}, and E.~Van~Andel.
\newblock Nyx: A massively parallel {AMR} code for computational cosmology.
\newblock {\em The Astrophysical Journal}, 765(1):39, 2013. \href{https://doi.org/10.1088/0004-637X/765/1/39}
{doi: {{%
10\hspace{.1pt}\discretionary{.}{%
}{.}\hspace{.4pt}1088\discretionary{/}{%
}{/}0004\discretionary{%
}{-}{-}637X\discretionary{/}{%
}{/}765\discretionary{/}{%
}{/}1\discretionary{/}{%
}{/}39}}}


\bibitem{Emilien-TMLR}
E.~Dupont, H.~Loya, M.~Alizadeh, A.~Golinski, Y.~W. Teh, and A.~Doucet.
\newblock {COIN++}: Neural compression across modalities.
\newblock {\em Transactions on Machine Learning Research}, 2022, 2022.

\bibitem{Finn-ICML17}
C.~Finn, P.~Abbeel, and S.~Levine.
\newblock Model-agnostic meta-learning for fast adaptation of deep networks.
\newblock In {\em Proceedings of International Conference on Machine Learning}, pp. 1126--1135, 2017. \href{https://doi.org/10.5555/3305381.3305498}
{doi: {{%
10\hspace{.1pt}\discretionary{.}{%
}{.}\hspace{.4pt}5555\discretionary{/}{%
}{/}3305381\hspace{.1pt}\discretionary{.}{%
}{.}\hspace{.4pt}3305498}}}


\bibitem{Gu-CG23}
P.~Gu, D.~Z. Chen, and C.~Wang.
\newblock {NeRVI}: Compressive neural representation of visualization images for communicating volume visualization results.
\newblock {\em Computers \& Graphics}, 116:216--227, 2023. \href{https://doi.org/10.1016/J.CAG.2023.08.024}
{doi: {{%
10\hspace{.1pt}\discretionary{.}{%
}{.}\hspace{.4pt}1016\discretionary{/}{%
}{/}J\hspace{.1pt}\discretionary{.}{%
}{.}\hspace{.4pt}CAG\hspace{.1pt}\discretionary{.}{%
}{.}\hspace{.4pt}2023\hspace{.1pt}\discretionary{.}{%
}{.}\hspace{.4pt}08\hspace{.1pt}\discretionary{.}{%
}{.}\hspace{.4pt}024}}}


\bibitem{Han-TVCG23}
J.~Han and C.~Wang.
\newblock {CoordNet}: Data generation and visualization generation for time-varying volumes via a coordinate-based neural network.
\newblock {\em IEEE Transactions on Visualization and Computer Graphics}, 29(12):4951--4963, 2023. \href{https://doi.org/10.1109/TVCG.2022.3197203}
{doi: {{%
10\hspace{.1pt}\discretionary{.}{%
}{.}\hspace{.4pt}1109\discretionary{/}{%
}{/}TVCG\hspace{.1pt}\discretionary{.}{%
}{.}\hspace{.4pt}2022\hspace{.1pt}\discretionary{.}{%
}{.}\hspace{.4pt}3197203}}}


\bibitem{Han-KD-INR}
J.~Han, H.~Zheng, and C.~Bi.
\newblock {KD-INR}: Time-varying volumetric data compression via knowledge distillation-based implicit neural representation.
\newblock {\em IEEE Transactions on Visualization and Computer Graphics}, 30(10):6826--6838, 2024. \href{https://doi.org/10.1109/TVCG.2023.3345373}
{doi: {{%
10\hspace{.1pt}\discretionary{.}{%
}{.}\hspace{.4pt}1109\discretionary{/}{%
}{/}TVCG\hspace{.1pt}\discretionary{.}{%
}{.}\hspace{.4pt}2023\hspace{.1pt}\discretionary{.}{%
}{.}\hspace{.4pt}3345373}}}


\bibitem{Hospedales-TPMAI22}
T.~M. Hospedales, A.~Antoniou, P.~Micaelli, and A.~J. Storkey.
\newblock Meta-learning in neural networks: A survey.
\newblock {\em IEEE Transactions on Pattern Analysis and Machine Intelligence}, 44(9):5149--5169, 2022. \href{https://doi.org/10.1109/TPAMI.2021.3079209}
{doi: {{%
10\hspace{.1pt}\discretionary{.}{%
}{.}\hspace{.4pt}1109\discretionary{/}{%
}{/}TPAMI\hspace{.1pt}\discretionary{.}{%
}{.}\hspace{.4pt}2021\hspace{.1pt}\discretionary{.}{%
}{.}\hspace{.4pt}3079209}}}


\bibitem{Li-TVCG24}
H.~Li and H.-W. Shen.
\newblock Improving efficiency of iso-surface extraction on implicit neural representations using uncertainty propagation.
\newblock {\em IEEE Transactions on Visualization and Computer Graphics}, 31(1):1--13, 2024. \href{https://doi.org/10.1109/TVCG.2024.3365089}
{doi: {{%
10\hspace{.1pt}\discretionary{.}{%
}{.}\hspace{.4pt}1109\discretionary{/}{%
}{/}TVCG\hspace{.1pt}\discretionary{.}{%
}{.}\hspace{.4pt}2024\hspace{.1pt}\discretionary{.}{%
}{.}\hspace{.4pt}3365089}}}


\bibitem{YF-Lu-VISSP24}
Y.~Lu, P.~Gu, and C.~Wang.
\newblock {FCNR}: Fast compressive neural representation of visualization images.
\newblock In {\em Proceedings of IEEE VIS Conference (Short Papers)}, pp. 31--35, 2024. \href{https://doi.org/10.1109/VIS55277.2024.00014}
{doi: {{%
10\hspace{.1pt}\discretionary{.}{%
}{.}\hspace{.4pt}1109\discretionary{/}{%
}{/}VIS55277\hspace{.1pt}\discretionary{.}{%
}{.}\hspace{.4pt}2024\hspace{.1pt}\discretionary{.}{%
}{.}\hspace{.4pt}00014}}}


\bibitem{Lu-neurcomp}
Y.~Lu, K.~Jiang, J.~A. Levine, and M.~Berger.
\newblock Compressive neural representations of volumetric scalar fields.
\newblock {\em Computer Graphics Forum}, 40(3):135--146, 2021. \href{https://doi.org/10.1111/cgf.14295}
{doi: {{%
10\hspace{.1pt}\discretionary{.}{%
}{.}\hspace{.4pt}1111\discretionary{/}{%
}{/}cgf\hspace{.1pt}\discretionary{.}{%
}{.}\hspace{.4pt}14295}}}


\bibitem{Nichol-arXiv18}
A.~Nichol, A.~Joshua, and J.~Schulman.
\newblock On first-order meta-learning algorithms.
\newblock {\em arXiv:1803.02999}, 2018. \href{https://doi.org/10.48550/arXiv.1803.02999}
{doi: {{%
10\hspace{.1pt}\discretionary{.}{%
}{.}\hspace{.4pt}48550\discretionary{/}{%
}{/}arXiv\hspace{.1pt}\discretionary{.}{%
}{.}\hspace{.4pt}1803\hspace{.1pt}\discretionary{.}{%
}{.}\hspace{.4pt}02999}}}


\bibitem{Popinet-JAOT04}
S.~Popinet, M.~Smith, and C.~Stevens.
\newblock Experimental and numerical study of the turbulence characteristics of airflow around a research vessel.
\newblock {\em Journal of Atmospheric and Oceanic Technology}, 21(10):1575--1589, 2004. \href{https://doi.org/10.1175/1520-0426(2004)021<1575:EANSOT>2.0.CO;2}
{doi: {{%
10\hspace{.1pt}\discretionary{.}{%
}{.}\hspace{.4pt}1175\discretionary{/}{%
}{/}1520\discretionary{%
}{-}{-}0426\discretionary{%
}{(}{(}2004\discretionary{)}{%
}{)}021{\textless}1575\discretionary{:}{%
}{:}EANSOT{\textgreater}2\hspace{.1pt}\discretionary{.}{%
}{.}\hspace{.4pt}0\hspace{.1pt}\discretionary{.}{%
}{.}\hspace{.4pt}CO\discretionary{;}{%
}{;}2}}}


\bibitem{Porter-VISSP19}
W.~P. Porter, Y.~Xing, B.~R. von Ohlen, J.~Han, and C.~Wang.
\newblock A deep learning approach to selecting representative time steps for time-varying multivariate data.
\newblock In {\em Proceedings of IEEE VIS Conference (Short Papers)}, pp. 131--135, 2019. \href{https://doi.org/10.1109/VISUAL.2019.8933759}
{doi: {{%
10\hspace{.1pt}\discretionary{.}{%
}{.}\hspace{.4pt}1109\discretionary{/}{%
}{/}VISUAL\hspace{.1pt}\discretionary{.}{%
}{.}\hspace{.4pt}2019\hspace{.1pt}\discretionary{.}{%
}{.}\hspace{.4pt}8933759}}}


\bibitem{Rojo-TVCG19}
I.~B. Rojo and T.~G{\"u}nther.
\newblock Vector field topology of time-dependent flows in a steady reference frame.
\newblock {\em IEEE Transactions on Visualization and Computer Graphics}, 26(1):280--290, 2019. \href{https://doi.org/10.1109/TVCG.2019.2934375}
{doi: {{%
10\hspace{.1pt}\discretionary{.}{%
}{.}\hspace{.4pt}1109\discretionary{/}{%
}{/}TVCG\hspace{.1pt}\discretionary{.}{%
}{.}\hspace{.4pt}2019\hspace{.1pt}\discretionary{.}{%
}{.}\hspace{.4pt}2934375}}}


\bibitem{silver1997tracking}
D.~Silver and X.~Wang.
\newblock Tracking and visualizing turbulent {3D} features.
\newblock {\em IEEE Transactions on Visualization and Computer Graphics}, 3(2):129--141, 1997. \href{https://doi.org/10.1109/2945.597796}
{doi: {{%
10\hspace{.1pt}\discretionary{.}{%
}{.}\hspace{.4pt}1109\discretionary{/}{%
}{/}2945\hspace{.1pt}\discretionary{.}{%
}{.}\hspace{.4pt}597796}}}


\bibitem{Sitzmann-MetaSDF-NeurIPS20}
V.~Sitzmann, E.~Chan, R.~Tucker, N.~Snavely, and G.~Wetzstein.
\newblock {MetaSDF}: Meta-learning signed distance functions.
\newblock In {\em Proceedings of Advances in Neural Information Processing Systems}, pp. 10136--10147, 2020. \href{https://doi.org/10.48550/arXiv.2006.09662}
{doi: {{%
10\hspace{.1pt}\discretionary{.}{%
}{.}\hspace{.4pt}48550\discretionary{/}{%
}{/}arXiv\hspace{.1pt}\discretionary{.}{%
}{.}\hspace{.4pt}2006\hspace{.1pt}\discretionary{.}{%
}{.}\hspace{.4pt}09662}}}


\bibitem{Sitzmann-SIREN-NeurIPS20}
V.~Sitzmann, J.~Martel, A.~Bergman, D.~Lindell, and G.~Wetzstein.
\newblock Implicit neural representations with periodic activation functions.
\newblock In {\em Proceedings of Advances in Neural Information Processing Systems}, pp. 7462--7473, 2020. \href{https://doi.org/10.48550/arXiv.2006.09661}
{doi: {{%
10\hspace{.1pt}\discretionary{.}{%
}{.}\hspace{.4pt}48550\discretionary{/}{%
}{/}arXiv\hspace{.1pt}\discretionary{.}{%
}{.}\hspace{.4pt}2006\hspace{.1pt}\discretionary{.}{%
}{.}\hspace{.4pt}09661}}}


\bibitem{Song-ACM23}
Y.~Song, T.~Wang, P.~Cai, S.~K. Mondal, and J.~P. Sahoo.
\newblock A comprehensive survey of few-shot learning: Evolution, applications, challenges, and opportunities.
\newblock {\em ACM Computing Surveys}, 55(13s):1--40, 2023. \href{https://doi.org/10.1145/3582688}
{doi: {{%
10\hspace{.1pt}\discretionary{.}{%
}{.}\hspace{.4pt}1145\discretionary{/}{%
}{/}3582688}}}


\bibitem{Tancik-CVPR21}
M.~Tancik, B.~Mildenhall, T.~Wang, D.~Schmidt, P.~P. Srinivasan, J.~T. Barron, and R.~Ng.
\newblock Learned initializations for optimizing coordinate-based neural representations.
\newblock In {\em Proceedings of IEEE Conference on Computer Vision and Pattern Recognition}, pp. 2846--2855, 2021. \href{https://doi.org/10.1109/CVPR46437.2021.00287}
{doi: {{%
10\hspace{.1pt}\discretionary{.}{%
}{.}\hspace{.4pt}1109\discretionary{/}{%
}{/}CVPR46437\hspace{.1pt}\discretionary{.}{%
}{.}\hspace{.4pt}2021\hspace{.1pt}\discretionary{.}{%
}{.}\hspace{.4pt}00287}}}


\bibitem{Tang-PVIS24}
K.~Tang and C.~Wang.
\newblock {ECNR}: Efficient compressive neural representation of time-varying volumetric datasets.
\newblock In {\em Proceedings of IEEE Pacific Visualization Conference}, pp. 72--81, 2024. \href{https://doi.org/10.1109/PacificVis60374.2024.00017}
{doi: {{%
10\hspace{.1pt}\discretionary{.}{%
}{.}\hspace{.4pt}1109\discretionary{/}{%
}{/}PacificVis60374\hspace{.1pt}\discretionary{.}{%
}{.}\hspace{.4pt}2024\hspace{.1pt}\discretionary{.}{%
}{.}\hspace{.4pt}00017}}}


\bibitem{Tang-CG24}
K.~Tang and C.~Wang.
\newblock {STSR-INR}: Spatiotemporal super-resolution for time-varying multivariate volumetric data via implicit neural representation.
\newblock {\em Computers \& Graphics}, 119:103874, 2024. \href{https://doi.org/10.1016/j.cag.2024.01.001}
{doi: {{%
10\hspace{.1pt}\discretionary{.}{%
}{.}\hspace{.4pt}1016\discretionary{/}{%
}{/}j\hspace{.1pt}\discretionary{.}{%
}{.}\hspace{.4pt}cag\hspace{.1pt}\discretionary{.}{%
}{.}\hspace{.4pt}2024\hspace{.1pt}\discretionary{.}{%
}{.}\hspace{.4pt}01\hspace{.1pt}\discretionary{.}{%
}{.}\hspace{.4pt}001}}}


\bibitem{Tang-VIS24}
K.~Tang and C.~Wang.
\newblock {StyleRF-VolVis}: Style transfer of neural radiance fields for expressive volume visualization.
\newblock {\em IEEE Transactions on Visualization and Computer Graphics}, 31(1):613--623, 2025. \href{https://doi.org/10.1109/TVCG.2024.3456342}
{doi: {{%
10\hspace{.1pt}\discretionary{.}{%
}{.}\hspace{.4pt}1109\discretionary{/}{%
}{/}TVCG\hspace{.1pt}\discretionary{.}{%
}{.}\hspace{.4pt}2024\hspace{.1pt}\discretionary{.}{%
}{.}\hspace{.4pt}3456342}}}


\bibitem{t-SNE}
L.~van~der Maaten and G.~Hinton.
\newblock Visualizing data using {t-SNE}.
\newblock {\em Journal of Machine Learning Research}, 9(86):2579–2605, 2008.

\bibitem{Wang-DL4SciVis}
C.~Wang and J.~Han.
\newblock {DL4SciVis}: A state-of-the-art survey on deep learning for scientific visualization.
\newblock {\em IEEE Transactions on Visualization and Computer Graphics}, 29(8):3714--3733, 2023. \href{https://doi.org/10.1109/TVCG.2022.3167896}
{doi: {{%
10\hspace{.1pt}\discretionary{.}{%
}{.}\hspace{.4pt}1109\discretionary{/}{%
}{/}TVCG\hspace{.1pt}\discretionary{.}{%
}{.}\hspace{.4pt}2022\hspace{.1pt}\discretionary{.}{%
}{.}\hspace{.4pt}3167896}}}


\bibitem{Weiss-CGF22}
S.~Weiss, P.~Herm{\"u}ller, and R.~Westermann.
\newblock Fast neural representations for direct volume rendering.
\newblock {\em Computer Graphics Forum}, 41(6):196--211, 2022. \href{https://doi.org/10.1111/cgf.14578}
{doi: {{%
10\hspace{.1pt}\discretionary{.}{%
}{.}\hspace{.4pt}1111\discretionary{/}{%
}{/}cgf\hspace{.1pt}\discretionary{.}{%
}{.}\hspace{.4pt}14578}}}


\bibitem{Whalen-TAJ08}
D.~Whalen and M.~L. Norman.
\newblock Ionization front instabilities in primordial {H II} regions.
\newblock {\em The Astrophysical Journal}, 673:664--675, 2008. \href{https://doi.org/10.1086/524400}
{doi: {{%
10\hspace{.1pt}\discretionary{.}{%
}{.}\hspace{.4pt}1086\discretionary{/}{%
}{/}524400}}}


\bibitem{Wurster-TVCG24}
S.~W. Wurster, T.~Xiong, H.-W. Shen, H.~Guo, and T.~Peterka.
\newblock Adaptively placed multi-grid scene representation networks for large-scale data visualization.
\newblock {\em IEEE Transactions on Visualization and Computer Graphics}, 30(1):965--974, 2024. \href{https://doi.org/10.1109/TVCG.2023.3327194}
{doi: {{%
10\hspace{.1pt}\discretionary{.}{%
}{.}\hspace{.4pt}1109\discretionary{/}{%
}{/}TVCG\hspace{.1pt}\discretionary{.}{%
}{.}\hspace{.4pt}2023\hspace{.1pt}\discretionary{.}{%
}{.}\hspace{.4pt}3327194}}}


\bibitem{Xiong-TVCG24}
T.~Xiong, S.~W. Wurster, H.~Guo, T.~Peterka, and H.-W. Shen.
\newblock Regularized multi-decoder ensemble for an error-aware scene representation network.
\newblock {\em IEEE Transactions on Visualization and Computer Graphics}, 31(1):645--655, 2025. \href{https://doi.org/10.1109/TVCG.2024.3456357}
{doi: {{%
10\hspace{.1pt}\discretionary{.}{%
}{.}\hspace{.4pt}1109\discretionary{/}{%
}{/}TVCG\hspace{.1pt}\discretionary{.}{%
}{.}\hspace{.4pt}2024\hspace{.1pt}\discretionary{.}{%
}{.}\hspace{.4pt}3456357}}}


\bibitem{Yao-PVIS25}
S.~Yao, Y.~Lu, and C.~Wang.
\newblock {ViSNeRF}: Efficient multidimensional neural radiance field representation for visualization synthesis of dynamic volumetric scenes.
\newblock In {\em Proceedings of IEEE Pacific Visualization Conference}, 2025.
\newblock Accepted.

\end{thebibliography}

\end{document}